\documentclass[sigconf,10pt]{acmart}
\usepackage{balance}
\usepackage[T1]{fontenc}
\usepackage{color,soul}
\soulregister\cite7
\soulregister\ref7
\soulregister\pageref7

\usepackage{background}

\backgroundsetup{angle=0, 
scale=1,
color=black,
firstpage=true,
position=current page.north,
hshift=0pt,
vshift=-20pt,
contents={\ifnum\value{page}=1 PREPRINT: Accepted at the 25th Annual International Conference on Mobile Computing and Networking (MobiCom), 2019 \else \fi}
}

\AtBeginDocument{%
\providecommand\BibTeX{{%
\normalfont B\kern-0.5em{\scshape i\kern-0.25em b}\kern-0.8em\TeX}}}

\title{MobiSR: Efficient On-Device Super-Resolution through Heterogeneous Mobile Processors}

\author{{Royson Lee$^\dagger$*, Stylianos I. Venieris$^\dagger$*\\ \L ukasz Dudziak$^\dagger$, Sourav Bhattacharya$^\dagger$, Nicholas D. Lane$^{\dagger,\ddagger}$}}

\authorwithoutinstuition{{Royson Lee, Stylianos I. Venieris, \L ukasz Dudziak, Sourav Bhattacharya, Nicholas D. Lane}}

\affiliation{\institution{$^\dagger$Samsung AI Center, Cambridge\hspace{+0.75cm}$^\ddagger$University of Oxford}{\Small\textit{{* Indicates equal contribution.}}}}

\renewcommand{\shortauthors}{Lee and Venieris, et al.}

\usepackage{natbib}
\usepackage{graphicx}
\usepackage{pbox}
\usepackage{balance}
\usepackage{multirow}
\usepackage[flushleft]{threeparttable}
\usepackage{afterpage}
\usepackage{subcaption}
\usepackage{lipsum}
\usepackage{tablefootnote}
\usepackage{booktabs}
\usepackage{extarrows}
\usepackage{stmaryrd}
\usepackage{gensymb}
\usepackage[ruled,linesnumbered]{algorithm2e}

\newlength\mylenin
\newcommand\myinput[1]{%
\settowidth\mylenin{\KwIn{}}%
\setlength\hangindent{\mylenin}%
\hspace*{\mylenin}#1\\}

\let\oldnl\nl 
\newcommand{\nonl}{\renewcommand{\nl}{\let\nl\oldnl}} 

\newlength\mylenout

\usepackage{amsmath}
\usepackage{amssymb}
\usepackage{dsfont}
\usepackage[bb=boondox]{mathalfa}

\begin{abstract}
In recent years, convolutional networks have demonstrated unprecedented performance in the image restoration task of super-resolution (SR). SR entails the upscaling of a single low-resolution image in order to meet application-specific image quality demands and plays a key role in mobile devices. To comply with privacy regulations and reduce the overhead of cloud computing, executing SR models locally on-device constitutes a key alternative approach. Nevertheless, the excessive compute and memory requirements of SR workloads pose a challenge in mapping SR networks on resource-constrained mobile platforms. This work presents MobiSR, a novel framework for performing efficient super-resolution on-device. Given a target mobile platform, the proposed framework considers popular model compression techniques and traverses the design space to reach the highest performing trade-off between image quality and processing speed. At run time, a novel scheduler dispatches incoming image patches to the appropriate model-engine pair based on the patch's estimated upscaling difficulty in order to meet the required image quality with minimum processing latency. 
Quantitative evaluation shows that the proposed framework yields on-device SR designs that achieve an average speedup of $2.13\times$ over highly-optimized parallel difficulty-unaware mappings 
and $4.79\times$ over highly-optimized single compute engine implementations.
\end{abstract}



\begin{document}

\copyrightyear{2019}
\acmYear{2019}
\acmConference[MobiCom '19]{The 25th Annual International Conference on Mobile Computing and Networking}{October 21--25, 2019}{Los Cabos, Mexico}
\acmBooktitle{The 25th Annual International Conference on Mobile Computing and Networking (MobiCom '19), October 21--25, 2019, Los Cabos, Mexico}
\acmPrice{15.00}
\acmDOI{10.1145/3300061.3345455}
\acmISBN{978-1-4503-6169-9/19/10}

\fancyhead{}

\maketitle

\section{Introduction}
The rapid progress of convolutional neural networks (CNNs) has led to substantial performance improvements in the computer vision task of super-resolution (SR). SR networks are capable of processing a low-resolution image and producing an output with a significant increase in resolution \citep{SRCNN}. This property has made CNN-powered SR an enabling technology for building novel applications on mobile and home devices, including mobile phones, electronic photograph frames and televisions. 

Despite their unparalleled performance, state-of-the-art SR networks \citep{EDSR,RCAN,RDN,IDN} pose significant deployment challenges. To upscale low-resolution images, SR models often propagate feature maps of large spatial dimensions across their layers, leading to an excessive number of operations and run-time storage requirements.

At the moment, to alleviate this computational barrier, service providers commonly employ cloud-computing solutions. Under this setup, an application collects frames and transmits them to a base server where powerful server-grade machines perform SR. However, in latency- and privacy-sensitive applications, the high response time and security risks of cloud computing may not be tolerable. Furthermore, the need for constant Internet connectivity and the power consumption overhead of exchanging data with the cloud together with the cost of hosting a data center often prohibits the offloading of computations. As a result, there is an emerging need to develop methods and systems that alleviate the limitations of cloud-based computing by executing SR networks using local on-device processing \cite{mcdnn_2016,nic2017embedded_dl, venieris2018deploying}. 

However, as SR networks are computationally expensive, achieving 30 fps using on-device resources is impractical for upscaling to large image resolutions. For instance, given that mobile digital cameras, such as Pixel 3's, are able to capture and stream in extremely high image resolutions, achieving such resolutions in real-time by running SR networks locally is currently unrealistic. Therefore, common realistic applications of SR on mobile, such as zoom, are image-centric, rather than video-focused. Another practical application of mobile SR involves saving data. Popular social media networks such as Facebook, Instagram and Reddit and messaging applications such as Snapchat are image-heavy applications which constantly use data as the user scrolls his feed or sends a message. Given the popularity of data-saving alternatives such as Facebook Lite, features that enable devices to download low-resolution images of a user's feed and/or messages and upscale them locally would be not only feasible, but also well-received. Moreover, minimizing the network bandwidth needed to load an image feed would allow the app to work more responsively under harsh network conditions and operate in areas with poor cloud connectivity.

In this paper, we propose MobiSR, a novel automated framework that pushes the performance of on-device SR networks. Drawing from the fact that not all inputs have the same upscaling difficulty, MobiSR introduces model compression as a design dimension for the local processing of SR models and introduces a hardware-aware scheduling scheme for allocating inputs to model-compute engine pairs. To explore the model space, the proposed framework starts from a user-supplied SR network and employs a set of compression techniques in order to generate multiple SR networks with varying accuracy-workload characteristics. Upon deployment, a difficulty evaluation unit estimates the upscaling difficulty of incoming samples. Based on the observation that some image patches are shown to be more difficult to upscale for both large and compact models, while some patches are handled better by larger models, the framework schedules inputs accordingly to strike an optimal balance between image quality and speed. Specifically, the inputs that are classified as difficult are computed using a less accurate, but compact model to obtain a rapid upscaling, while easier inputs are assigned to a larger, but more accurate model. Overall, \mbox{MobiSR} considers the error tolerance of the target application in order to perform model selection and tailors its scheduling policy to both the selected SR models and the available compute engines. The key contributions of this paper are the following:

\begin{itemize}
\item The introduction of a two-model super-resolution system that exploits the upscaling difficulty of incoming patches to boost the performance of on-device SR. A novel tunable difficulty evaluation unit is presented that estimates the upscaling difficulty of incoming image patches and schedules them across different model-compute engine pairs at run time.
\item A design space exploration methodology that considers the user-supplied SR model and the target mobile platform together with a user-specified error tolerance and generates an optimized SR system. By treating model compression as a design dimension and employing a hardware-aware scheduling policy, the proposed methodology explores candidate designs at both the model and scheduling level and generates an SR system tailored to meet the user-specified error tolerance at the minimum latency.
\end{itemize}

\section{Background}
\label{sec:background}
Since the introduction of using CNNs for SR tasks in \cite{SRCNN}, there has been a surge in SR models that utilized popular techniques such as attention \cite{Bahdanau_2014}, residual blocks \cite{He_2016}, and generative adversarial networks \cite{Goodfellow_2014}. These models aim to either map low-resolution images closer to their high-resolution ground truth or make SR images look more naturally pleasing. The former, which are usually trained on either the L1 or Mean Square Error (MSE) loss, favour pixel-to-pixel comparisons and are evaluated on image distortion metrics such as MSE, Peak Signal-to-Noise Ratio (PSNR), and Structural Similarity Index (SSIM) \cite{SSIM}. The latter, on the other hand, are usually trained using a combination of different loss functions, including perceptual \cite{Johnson_2016} and adversarial losses \cite{Goodfellow_2014}. These models focus on the perceptual quality of the image and are evaluated on no-reference metrics such as Natural Image Quality Evaluator \cite{NIQE} and perceptual score \cite{Ma_2017}. In this work, we focus on the former, \textit{i.e.} mapping low-resolution images closer to their high-resolution ground truth.

Unlike models that are optimized for discriminative tasks, SR models are resource-intensive networks as each layer needs to maintain or upscale the spatial dimensions of its feature maps. As a result, the number of multiply-add operations are typically counted in the billions as opposed to millions in discriminative networks \cite{embench_2019}. Although the research community has made a few steps towards constructing efficient SR models that are optimized for mobile platforms, \mbox{(1) running} these models on-device is still costly and (2) popular compression techniques have not yet been utilized to derive lightweight, mobile-friendly variants. For instance, in the experiments presented in Section \ref{sec:eval}, the winning model \cite{Vu2018FastAE} of the recent 2018 PIRM Challenge on perceptual SR on mobile \cite{Ignatov_2018} requires more than 1.4 s to $\times$4 upscale an image to 720p on the Hexagon DSP of Qualcomm Snapdragon 845.

\begin{figure}[t]
\centering
\includegraphics[trim={4cm 3cm 3cm 1cm},clip,width=0.5\textwidth]{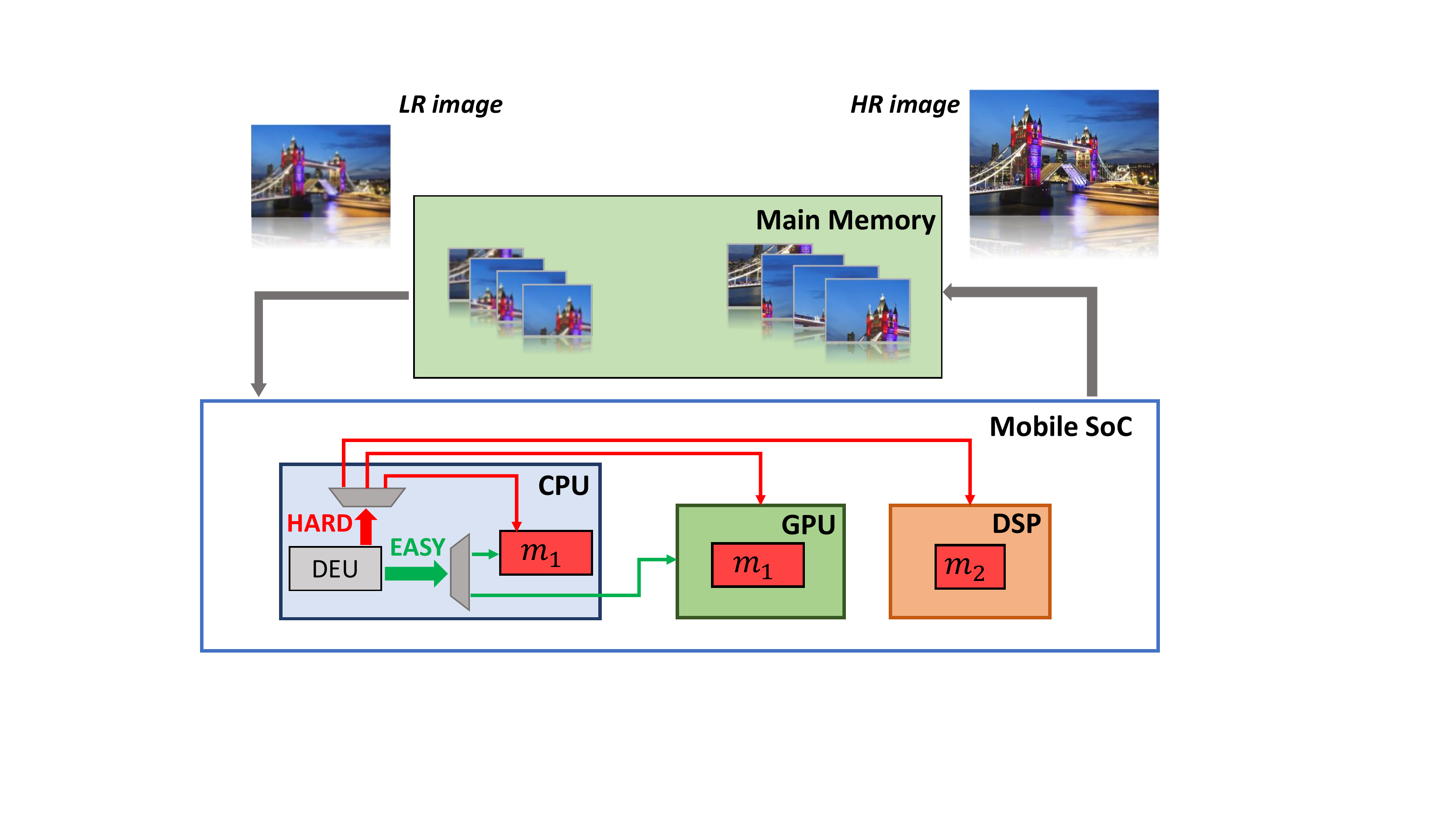}
\vspace{-0.8cm}
\caption{MobiSR's system architecture.}
\label{fig:sys_arch}
\vspace{-0.5cm}
\end{figure}

\textbf{Challenges of on-device SR.}
As the size of the image increases, running SR models on a single compute engine is difficult to scale or even impractical; upscaling a large image may lead to a memory overload. Cloud-based solutions \cite{Caulfield2016,Fowers_2018,Hazelwood_2018} can be deployed to offload the expensive computation. However, such solutions rely on a fast and stable communication channel and the need to maintain privacy, assumptions that are difficult to achieve in practice. 
Another solution would be to load-balance the computation of upscaling across the available on-device compute engines of the target mobile System-on-Chip (SoC). However, naively load-balancing SR models on multiple compute engines fails to utilize hardware-specific optimizations; different compute engines are optimized for different types of layers of a network. Furthermore, using reduced-precision compute engines in an uninformed manner can substantially affect the application-level quality of result (QoR). Therefore, there is a need to better utilize on-device resources to improve both the efficiency and scalability of running SR models locally.

\textbf{SR model compression.}
So far, substantial effort has been invested on developing network compression techniques, such as pruning \cite{Han_2015,Yang_2017_CVPR}, quantization \cite{Han_2016}, and knowledge distillation \cite{Hinton_2015}, for building efficient neural networks. In particular, a number of convolution approximations, such as low-rank tensor decomposition \cite{Sifre_2014}, have been successfully employed as a primary component in building fast and accurate discriminative vision models \cite{mobilenetv1,shufflenetv1,clcnet,effnet}. These techniques typically aim to express a convolution as a sequence of simpler tensor operations, reducing in this manner the storage and computation cost of the network. With current SR models being excessively large, exploiting the potential of existing compression techniques can lead to significant gains in efficiency. Nevertheless, each technique provides varying gains depending on the target hardware optimizations, but also on the model and the level of quantization involved. Therefore, a key challenge in accelerating SR models is selecting appropriate convolution-approximation techniques based on both their impact on the accuracy of the given model and their efficient mapping on the available compute engines.

\begin{figure}[t]
\centering
\includegraphics[trim={0cm 2cm 0cm 1cm},clip,width=0.5\textwidth]{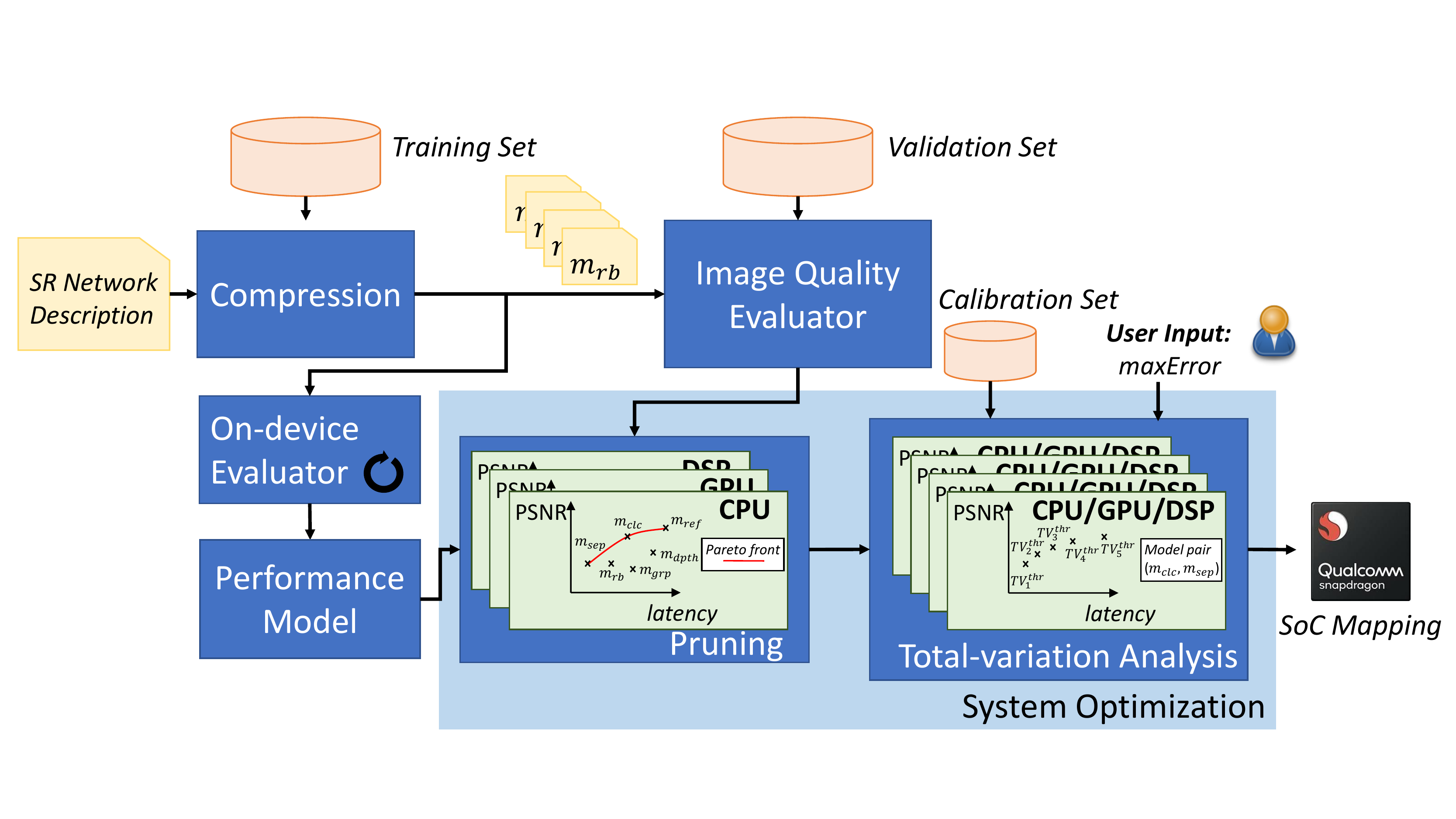}
\vspace{-0.6cm}
\caption{MobiSR's processing flow.}
\label{fig:detailed_flow}
\vspace{-0.7cm}
\end{figure}

\section{M\lowercase{obi}SR}
In this section, we present the high-level flow of MobiSR followed by a detailed description of its internal components.

\subsection{Overview}
\label{sec:framework}

Given a particular SR task, MobiSR searches the space of candidate on-device designs and generates a two-model system optimized for the target mobile platform. Upon deployment, the generated super-resolution system (Fig. \ref{fig:sys_arch}) consists of:

\begin{itemize}
\item A compact network mapped on a high-performance compute engine that trades-off QoR with low processing latency; this can be an aggressively compressed model running on the DSP of the target mobile SoC.
\item A large network which guarantees the user-specified QoR at the cost of a larger workload; this can be a lightly compressed model or a user-defined reference model running on the CPU and GPU of the target SoC.
\item A tunable difficulty-aware scheduler that parallelizes the incoming low-resolution image by dispatching each image patch to the appropriate model-compute engine pair based on its estimated \textit{upscaling difficulty}. 

\end{itemize}
The key idea behind the proposed approach is that, instead of processing the full set of patches using the large and expensive network, inputs that are classified as hard-to-upscale for both networks are rapidly processed by the compact network, with only a fraction of the inputs processed by the expensive large network, reducing in this way the overall latency of the system. Furthermore, the distortion that is induced due to the network compression of the compact model is restored by tuning the portion of images processed by each network based on the user-specified error threshold.

A high-level overview of MobiSR's flow is presented in \mbox{Fig. \ref{fig:detailed_flow}}. The framework is supplied with a high-level description of an SR network (\textit{i.e.} \mbox{PyTorch}\footnote{https://pytorch.org/} model), the specifications of the target mobile platform and an error tolerance in an image reconstruction quality metric (\textit{e.g.} PSNR). As a first step, the \textit{Compression} module applies a set of transformations over the supplied network in order to modify its topology and generate a number of compressed variants. To characterize their latency-QoR trade-off, each model is evaluated with respect to both its SR performance and on-device processing latency by the \textit{Image Quality} and \textit{On-device Evaluator} respectively. The \textit{On-device Evaluator} performs a number of runs on the compute engines of the target mobile platform and measures the average latency of each (model, compute engine) pair. 

Given the latency measurements, an analytical performance model is populated which enables the rapid estimation of the attainable latency for different scheduling schemes across the available devices. 
Next, the \textit{Pruning} module takes as input the (PSNR, latency) of each (model, compute engine) pair as generated by the \textit{Image Quality} and \textit{On-device Evaluators}. By examining the PSNR-latency space of each compute engine, only the models that lie on the Pareto front are kept, with the rest of the dominated models discarded as inefficient, reducing in this manner the space of candidate models. After the pruning step, the \textit{Total-variation Analysis} module is responsible for both tuning the difficulty-aware scheduler and selecting the models to be mapped on the target platform. Overall, given the user-specifed error tolerance, MobiSR generates a two-model system together with an associated scheduler tailored for the target mobile platform. 

\subsection{Model Space}
\label{subsec:model_space}
In MobiSR, the user-supplied SR model comprises the starting point for model selection. In this setting, the space of candidate models is determined by the techniques employed by our framework in order to modify the topology of the reference network. The complete model space is formed by defining a set of model transformations to change the complexity-QoR characteristics of the reference model.  
Given the computation cost of a standard $K_h$$\times$$K_w$ convolution
\begin{equation}
\label{eq:standard_conv}
S \cdot D \cdot K_h \cdot K_w \cdot F_h \cdot F_w 
\end{equation}
where $S$ is the number of input channels, $D$ is the number of output channels and $F_h$$\times$$F_w$ is the feature map size, MobiSR employs the following set of transformations:

\textbf{Residual Bottleneck Block $rb(r)$:} First introduced in the ResNet model \cite{He_2016}, the residual bottleneck design substitutes a conventional convolutional layer with a 1$\times$1 convolutional layer, used to compress the number of channels by a reduction factor $r$, followed by a $K_h \times K_w$ convolutional layer. Then, another 1$\times$1 convolutional layer along with a skip connection are employed to recover the number of output channels. The reduction in computation cost over a standard $K_h \times K_w$ convolutional layer is therefore
\begin{equation}
\label{eq:bn_technique}
\frac{S}{D \cdot K_h \cdot K_w \cdot r} + \frac{S}{r^2 \cdot D} + \frac{1}{K_h \cdot K_w \cdot r}
\end{equation}

\textbf{Group Convolutions $grp(g)$:}
The use of group convolutions \cite{Krizhevsky_2012} was introduced as a method of reducing the number of both parameters and operations with minimal impact on task-level performance \cite{Xie_2017}. This is achieved by splitting the convolutions channel-wise and computing them separately. In other words, the input feature maps are grouped and convolution is performed independently in each group. This leads to a computation cost reduction of $\frac{1}{g}$ as compared to a standard convolutional layer.

\textbf{Depthwise Separable Convolutions $dpth$:}
Depthwise convolutions are referred to as group convolutions in which the number of input channels is equal to the number of groups, $S = g$. In order for information to flow among groups, depthwise convolutions are usually paired with a 1$\times$1 convolution and the combination is known as depthwise separable convolution, which was first introduced in \cite{Sifre_2014} and termed in \cite{mobilenetv1}. From a workload perspective, depthwise separable convolutions yield a computation cost reduction of
\begin{equation}
\label{eq:dpth_conv}
\frac{1}{D} + \frac{1}{(K_h \cdot K_w)^2}
\end{equation}

\textbf{Separable Convolutions $sep$:}
This technique substitutes each $K_h \times K_w$ convolutional layer with a 1$\times K_h$ followed by a $K_w \times$1 convolution, separating the convolution dimension-wise and resulting in a computation cost reduction of  
\begin{equation}
\label{eq:sep_conv}
\frac{1}{K_h} + \frac{1}{K_w}
\end{equation}

\textbf{Inverted Residual Blocks $invr(e)$:}
Inverted residual blocks expand the number of channels by an expansion factor of $e$ by means of a 1$\times$1 convolution, followed by a $K_h \times K_w$ convolution and another 1$\times$1 convolution to recover the initial number of channels. This technique enables the use of skip connections directly on the bottleneck layers, resulting in an increase in computation cost, which is equal to that of Eq. (\ref{eq:bn_technique}) with $r = \frac{1}{e}$, but also in performance. Due to the increase in workload, inverted residual blocks were used together with depthwise convolutions when first introduced in \cite{mobilenetv2}.

\textbf{Channel Shuffle $chlshf$:}
Channel shuffling was introduced in \cite{shufflenetv1} to improve representational capability by changing the order of the channels, allowing information flow among channel groups. Specifically, an output of a grouped convolutional layer, which has $g$ groups of $\frac{D}{g}$ channels each, is reshaped into $\left(g, \frac{D}{g}\right)$, transposed into $\left(\frac{D}{g}, g\right)$, and flattened back to the number of output channels, $D$.

\textbf{Channel Split $chlsplt$:}
The splitting of feature channels into branches is termed as a "channel split" in \cite{shufflenetv2} and was introduced to improve processing speed. For instance, \cite{shufflenetv2} uses channel splitting to split the number of channels into two branches. Convolutions are performed only on a single branch before both branches are concatenated, resulting in a reduction in workload.

Given these compression methods, we define the transformations set $T$ as follows:
\begin{equation}
\label{eq:transf_set}
T = \left\{ rb(r), grp(g), dpth, sep, invr(e), chlshf, chlsplt \right\}
\end{equation}
To generate a new candidate model, we apply one transformation from the transformations set over the reference model:
\begin{equation}
\label{eq:model_build}
m \xlongleftarrow{t} m_{\text{ref}}, ~~~ t \in T 
\end{equation}
Formally, we capture the configuration of a model by defining a tuple representation of $m$
and the overall model space by means of a model set $\mathcal{M}$ (Eq. (\ref{eq:model_set})) that contains all reachable candidate models.

\begin{equation}
\label{eq:model_set}
\mathcal{M} = \left\{ m ~|~ m = \left< m_{\text{ref}}, T^*, \boldsymbol{\theta} \right>  \right\}, ~~~ T^* \subset T
\end{equation}
where $m_{\text{ref}}$ is the topology of the reference model, $T^*$ is the subset of applied transformations that are applied on $m_{\text{ref}}$ to obtain $m$, and $\boldsymbol{\theta}$ are the learned parameters of $m$ after the training process.

\begin{figure}[t]
\vspace{-0.75cm}
\centering
\includegraphics[width=0.5\textwidth]{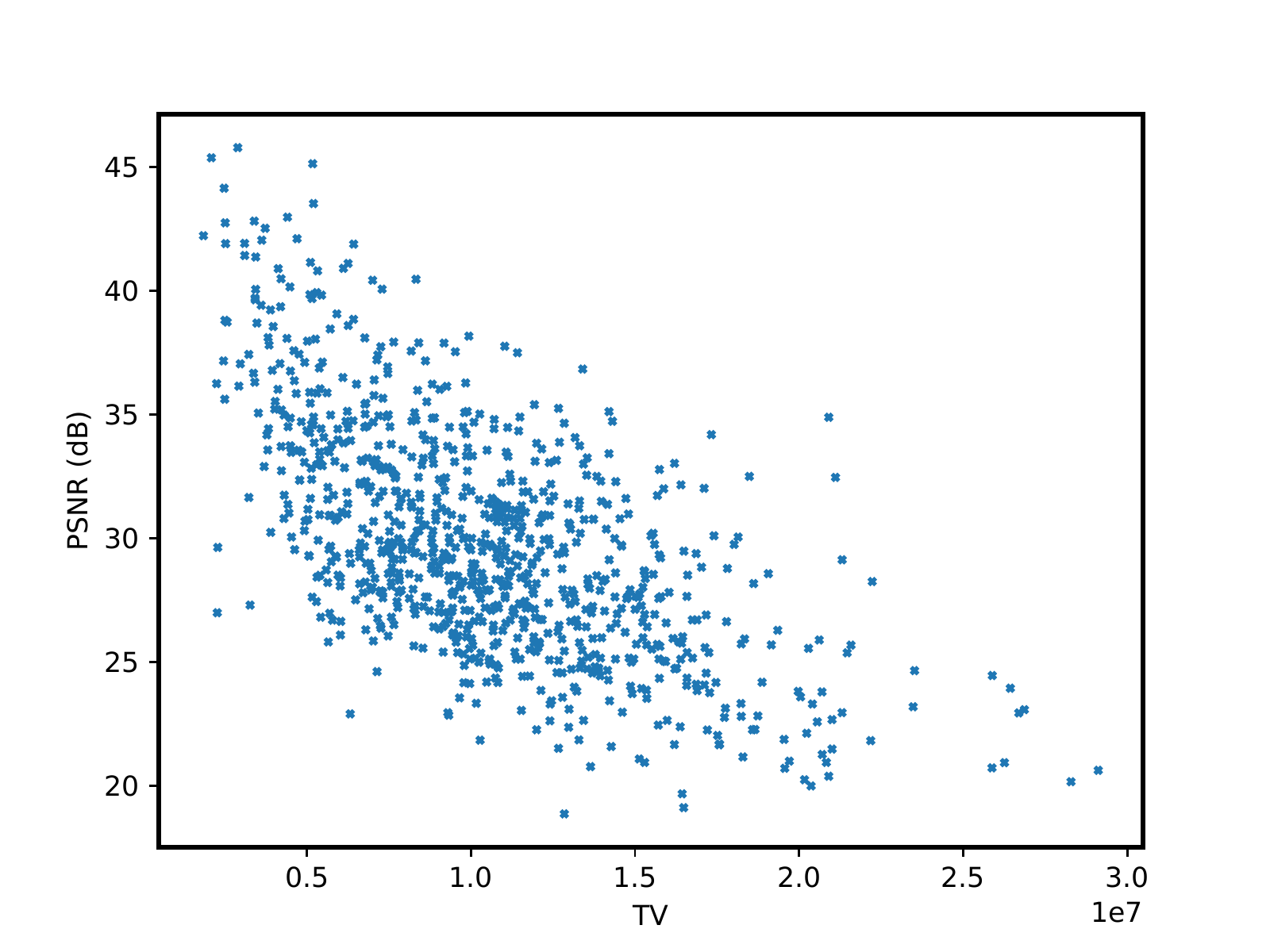}
\caption{Low-resolution images along with their TV and the PSNR achieved after $\times$4 upscaling using our reference model, $m_{\text{ref}}$, for images in the DIV2K training and validation dataset.}
\label{fig:psnr_tv}
\end{figure}

\begin{figure}[t]
\vspace{-0.7cm}
\centering
\includegraphics[width=0.5\textwidth]{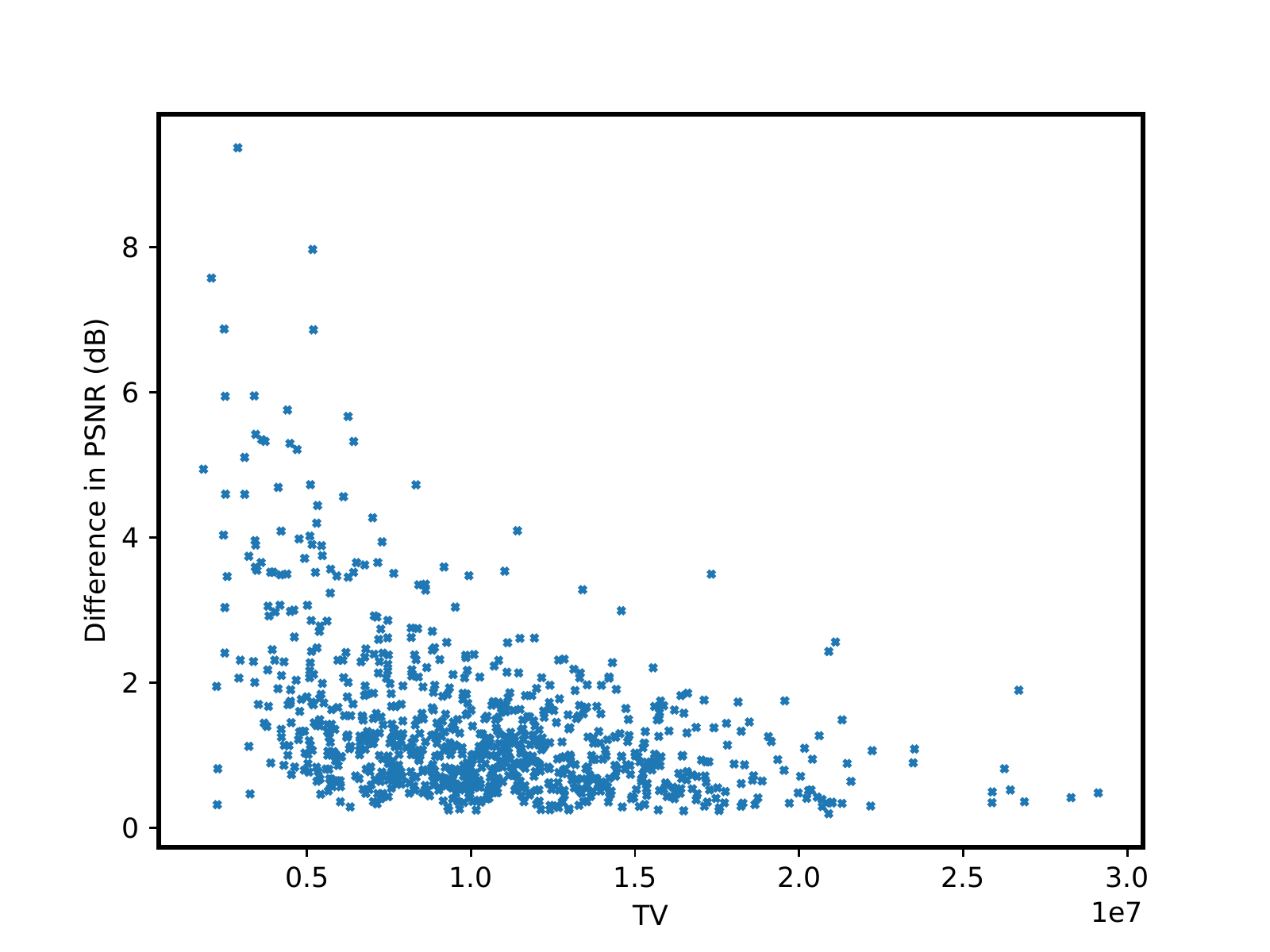}
\caption{PSNR difference of $\times$4 upscaling between our reference model, $m_{\text{ref}}$, and a more compact model, $m_{\text{s2}}$.}
\label{fig:psnr_diff_tv}
\end{figure}

\subsection{Difficulty Evaluation Unit}
\label{subsec:deu}
To sustain the QoR within the tolerance bounds of the user while achieving higher processing speed, MobiSR exploits the fact that not all image patches have the same upscaling difficulty. To this end, the Difficulty Evaluation Unit (DEU) is responsible for examining each patch and determining its complexity. To estimate upscaling difficulty, we employ the total variation (TV) metric \cite{total_variation_1992}. Total variation captures the complexity of an image by examining its spatial variation, with its anistropic version for a patch $p$ defined as:
\begin{equation}
\label{eq:total_var}
TV(p) = \sum_{i,j} |p_{i+1,j}-p_{i,j}| + |p_{i,j+1}-p_{i,j}|
\end{equation}

\begin{figure}[t!]
\centering
\captionsetup[subfigure]{justification=centering}
\begin{subfigure}[t]{0.25\textwidth}
    \centering
    \includegraphics[height=1.0in]{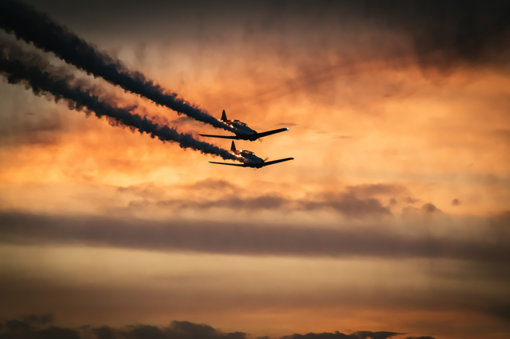}
    \caption{0301.png \\ PSNR: 45.38 / TV: 0.2e7}
    \label{fig:low_tv}
\end{subfigure}%
~
\begin{subfigure}[t]{0.25\textwidth}
    \centering
    \includegraphics[height=1.0in]{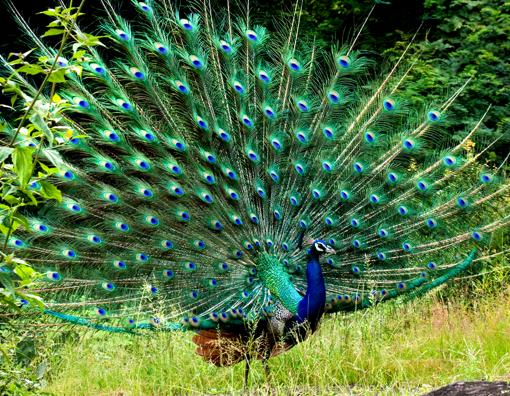}
    \caption{0063.png \\ PSNR: 20.62 / TV: 2.9e7}
    \label{fig:high_tv}
\end{subfigure}
        \caption{TV of low-resolution images from the DIV2K dataset and the PSNR achieved after $\times$4 upscaling using our reference model, $m_{\text{ref}}$. }
\label{fig:visual_psnr_tv}
\end{figure}

Fig. \ref{fig:visual_psnr_tv} presents a visual comparison between two images with low (Fig. \ref{fig:low_tv}) and high (Fig. \ref{fig:high_tv}) TV values together with the associated PSNR achieved after $\times$4 upscaling using our reference model, $m_{\text{ref}}$. As illustrated in the figures, an image consisting of unstructured fine details and texture (Fig. \ref{fig:high_tv}), has a higher TV and is harder to upscale compared to a highly structured or smoother image (Fig. \ref{fig:low_tv}).

To investigate the relationship between upscaling difficulty and TV of a given image in super-resolution settings, we examined the TV of each image in the DIV2K training and validation sets, together with the achieved PSNR obtained by our reference model, $m_{\text{ref}}$, which is described in \mbox{Section \ref{sec:eval}}. As depicted in Fig. \ref{fig:psnr_tv}, images with higher TV values tend to yield lower PSNR and hence are harder to upscale, while lower-TV images tend to reach higher PSNR and thus are upscaled with higher quality.

Following these observations, we define an image patch as hard-to-upscale based on the following criterion:
\begin{equation}
\label{eq:tv_criterion}
TV(p) > TV^{\text{thr}}
\end{equation}
where the TV threshold $TV^{\text{thr}}$ is a tunable parameter whose value is automatically configured by MobiSR as discussed in Section \ref{subsec:opt}.

\textbf{Upscaling-Difficulty-aware Scheduling.}
After computing the TV of an incoming patch, the DEU is responsible for dispatching it to the suitable model between $m_1$ and $m_2$. The goal is to employ a scheduling strategy that will not exceed the user-specified error tolerance and will yield the lowest latency. To this end, we explore the behavior of the model pair ($m_\text{ref}$, $m_\text{s2}$) on patches with varying TV values. Fig. \ref{fig:psnr_diff_tv} shows the PSNR difference between models $m_\text{ref}$ and $m_\text{s2}$ as a function of the value of TV for the DIV2K training and validation set. As observed from the figure, patches that are harder to upscale based on the TV criterion (\textit{i.e.} towards the right in Fig. \ref{fig:psnr_diff_tv}) are almost equally hard for the two models. On the other hand, on easier-to-upscale patches, the larger model is able to achieve significantly higher PSNR. To exploit this property, an upscaling-difficulty-aware scheduling policy is proposed which directs easy-to-upscale patches to the larger model and hard-to-upscale patches to the more compact model. In this manner, higher-TV patches that are almost equally hard for both models are processed rapidly using the more compact model, with easier patches processed by the larger model to sustain the PSNR within the specified bounds.

Algorithm \ref{alg:load_balance} presents the overall scheduling scheme. Model $m_1$ is mapped on the CPU and GPU engines with model $m_2$ mapped on the available DSP. Instead of solely using the per-patch upscaling difficulty as a scheduling criterion, load balancing is also employed to sustain the utilization of the available compute engines high. In this setting, Algorithm \ref{alg:load_balance} takes as inputs the SR model pair ($m_1, m_2$), the estimated execution time $t_{ce}$ for processing a patch with model $m$ on compute engine $ce$ and the selected TV threshold $TV^{\text{thr}}$. For each patch, the DEU first computes the associated TV value (line 3) and then dispatches the patch to the appropriate model-compute engine pair based on the total-variation criterion (line 4). In the case of an easy-to-upscale patch, the patch is allowed to be processed only by $m_1$ and thus the DEU dispatches the patch to either the CPU or the GPU, aiming to balance the load of the two engines (lines 5-7). In the case of hard-to-scale patches, the DEU allows the patch to be directed to $m_2$, but also includes $m_1$ as a candidate in order to avoid oversubscription of $m_2$'s compute engine. Since processing a patch with $m_1$ does not degrade the resulting PSNR, hard patches are also allowed to be processed by $m_1$ in case the DSP is overloaded (lines 8-10). On the other hand, easy patches are restricted to run using $m_1$ in order to avoid a significant quality loss due to $m_2$'s compression.

The range of values of total variation tends to vary between different domains. To estimate the dynamic range of TV on the target domain, MobiSR employs a user-supplied calibration set consisting of a small number of input samples. Given a few patches, the dynamic range of TV for a given dataset is estimated in order to tune the domain-specific total-variation threshold, $TV^{\text{thr}}$. 

\setlength{\textfloatsep}{0pt}
\begin{algorithm}[!t]
\footnotesize
\SetAlgoLined
\LinesNumbered
\DontPrintSemicolon
\KwIn{Image $I$}
\nonl
\myinput{SR model $m$}
\nonl
\myinput{Execution time per compute engine $\boldsymbol{t}^{\text{CE}}_{\{\mathcal{CE}\}} \in \mathbb{R}^{+|\mathcal{CE}|}_0$}
\nonl
\myinput{Total-variation threshold $TV^{\text{thr}}$}
$\boldsymbol{t}^{\text{end}}_{\{\mathcal{CE}\}} = \mathbb{0}^{|\mathcal{CE}|}$ // \textit{End time per compute engine}\;
\ForEach{patch $p \in I$}{
    $TV \leftarrow$ CalcTV($p$)\; 
    \uIf{$TV \le TV^{\text{thr}}$}{
        $i \leftarrow \text{argmin}~ \Big( \boldsymbol{t}^{\text{end}}_{\{\text{CPU,GPU}\}}$
        \\
        %
        \hspace*{11em}
        \rlap{\smash{$\left.\begin{array}{@{}c@{}}\\{}\\{}\\{}\\{}\end{array}\color{black}\right\}%
        \color{black}\begin{tabular}{l}PSNR-preserving engines\\ (\textit{e.g.} CPU, GPU) w/ restricted \\ load balancing.\end{tabular}$}}\;
        $\mbox{}\phantom{i \leftarrow \text{argmin}~ } + \boldsymbol{t}^{\text{CE}}_{\{\text{CPU,GPU}\}} \Big)$\;
    }
    \Else{
        $i \leftarrow \text{argmin}~ \left( \boldsymbol{t}^{\text{end}} + \boldsymbol{t}^{\text{CE}} \right)$\;
        \hspace*{11em}
        \rlap{\smash{$\left.\begin{array}{@{}c@{}}\\{}\\{}\\{}\\{}\end{array}\color{black}\right\}%
        \color{black}\begin{tabular}{l}Low-precision engines \\(\textit{e.g.} DSP) w/ load balancing\\  across $\mathcal{CE}$.\end{tabular}$}}\;
    }
    $\boldsymbol{t}^{\text{end}}_i \leftarrow \boldsymbol{t}^{\text{end}}_i + t_{i}^{\text{CE}}$\;
}
\caption{\small Upscaling-difficulty-aware scheduling for \mbox{parallel} load-balanced on-device super-resolution}
\label{alg:load_balance}
\end{algorithm}

\subsection{Performance Model}
\label{subsec:perf_model}

To efficiently explore different candidate designs without the need for implementations, a performance model is constructed that rapidly estimates a design's latency. 
To formally capture the processing resources of the target mobile platform, we define a compute engine set, $\mathcal{CE}$, which includes the compute engines that are available on the target chipset. 
In general, $\mathcal{CE}$ can represent a diversity of mobile SoCs hosting heterogeneous compute engines, ranging from the ubiquitous mobile CPUs and GPUs to the newer emerging NPUs \cite{ai_benchmark_2018}.
For instance, Qualcomm Snapdragon 845 SoC (SDM845) is represented as $\mathcal{CE}_{\text{SDM845}} = \left\{ CPU, GPU, DSP \right\}$. With this formulation, given an SR model $m$ and a single compute engine $ce \in \mathcal{CE}$, the execution time of upscaling an image $I$ using the $(m,ce)$ pair is estimated as:
\begin{equation}
\label{eq:ce_exec_time_bl}
t_{ce}^{\text{total}}(I, m) = \sum_{\text{patch}~ p \in I} t_{ce}(p, m)
\vspace{-0.2cm}
\end{equation}
where $t_{ce}(p,m)$ is the execution time for a single patch $p$ when model $m$ is mapped on compute engine $ce$. The per-patch execution time $t_{ce}(p,m)$ is measured by the \textit{On-device Evaluator} by means of a number of benchmark runs.

Following our difficulty-aware scheduling presented in Section \ref{subsec:deu}, each model-compute engine pair processes only the samples that lie within its total-variation threshold, $TV^{\text{thr}}$. To capture this strategy the execution time model is modified as follows:
\begin{small}
\label{eq:ce_exec_time_tv}
\[t_{ce}^{\text{total}}(I, m, TV^{\text{thr}}) \text{=}
    \begin{cases} 
        \sum_{p \in I} t_{ce}(p, m) {\mathds{1}(TV(p) \le TV^{\text{thr}})}, & m ~\text{is}~ m_1 \\ 
        \sum_{p \in I} t_{ce}(p, m) {\mathds{1}(TV(p) > TV^{\text{thr}})}, & m ~\text{is}~ m_2
    \end{cases}\]
    \vspace{0.15cm}
\end{small}

\noindent
where $\mathds{1}(\cdot)$ is the unity function that evaluates to $1$ when its bracketed condition is true. MobiSR distributes patches across the available engines in order to maximize the utilization of the on-chip compute resources and exploit the inherent parallelism across independent patches. Under this scheme, the overall latency of upscaling image $I$ using model $m$ on the target SoC is estimated as in Eq. (\ref{eq:latency}).
\begin{small}
\begin{align}
    \label{eq:latency}
    L(I, m, TV^{\text{thr}}, \mathcal{CE}) &= \max\left( \left\{ t_{ce}^{\text{total}}(I, m, TV^{\text{thr}}) ~~|~~ ce \in \mathcal{CE} \right\} \right) \\ 
    &+ t^{\text{stitch}} \nonumber 
\end{align}
\vspace{-0.5cm}
\end{small}

\noindent
where the first term captures the parallel execution of patches across engines and $t^{\text{stitch}}$ represents the overhead of assembling together the partial results of all patches to form the final high-resolution image.

\subsection{System Optimization}
\label{subsec:opt}
The developed framework aims to determine a pair of models together with a total-variation threshold that minimize the processing latency of performing on-device SR on the target mobile platform, given a user-supplied error tolerance. In this context, we pose the following optimization problem:
\begin{align}
\label{eq:obj_func}
&\min_{(m_1, m_2) , TV^{\text{thr}}}~L\left(I, (m_{1}, m_{2}), TV^{\text{thr}}, \mathcal{CE} \right) \\ 
&\text{ s.t. } \text{PSNR}(I, m_{\text{ref}}, 0) - \text{PSNR}\left( I, (m_{1}, m_{2}), TV^{\text{thr}} \right) \le \epsilon_{\text{max}}
\nonumber
\end{align}
where $L$, $TV^{\text{thr}}$ and $\epsilon_{\text{max}}$ are the latency in s/input, the total-variation threshold and the user-specified error tolerance respectively. Under this formulation, the objective function aims to find the tuple $\left<(m_1, m_2), TV^{\text{thr}} \right>$ that minimizes latency with a constraint on the degradation of QoR as captured by PSNR.

Given a reference SR model $m_{\text{ref}}$, the optimization problem in Eq. (\ref{eq:obj_func}) is defined over all candidate model pairs in the model space $\mathcal{M}$ presented in Section \ref{subsec:model_space}. Formally, we express this as the product $\mathcal{M} \times \mathcal{M}$. Furthermore, each pair can be deployed with a different TV threshold and therefore, given $N_{\text{TV}}$ discrete candidate total-variation thresholds, the total number of alternative designs to be explored can be calculated as follows:
\begin{equation}
\label{eq:designs_total}
|\mathcal{M}|^2 \cdot N_{\text{TV}}
\end{equation}
In this setup, the objective function $L:\{\mathcal{M} \times \mathcal{M},TV^{\text{thr}} \}$$\rightarrow$$\mathbb{R}^+$ can be evaluated for all $(m_1, m_2) \in \mathcal{M} \times \mathcal{M}$ by means of the performance model of Section \ref{subsec:perf_model}. In theory, the optimal design could be obtained by means of an exhaustive search with complete enumeration of all possible designs.

With latency and PSNR being a function of the TV of each patch of the processed image, evaluating $L(\cdot)$ and $\text{PSNR}(\cdot)$ is data-dependent and hence requires running each possible design $\left<(m_1, m_2), TV^{\text{thr}} \right>$ over a task-specific dataset to assess its attainable PSNR and latency. To avoid the overhead of an excessive number of evaluation runs, MobiSR employs two strategies for pruning the design space: 1) for each compute engine, we keep only the models that lie on the Pareto front of the PSNR-latency space. In this manner, models that are dominated with respect to their PSNR-latency balance on a given compute engine are discarded as inefficient; and 2) we impose the constraint that $m_2$ is equally or more compact than $m_1$. In this manner, we guide MobiSR to select two models with different PSNR-latency characteristics, in order to combine the high PSNR of $m_1$ with the fast processing of $m_2$.

After the pruning stage, MobiSR searches the remaining design space to determine the highest performing configuration of the tuple $\left< (m_1, m_2), TV^{\text{thr}} \right>$. To enable fast and exhaustive exploration, the developed performance model of Section \ref{subsec:perf_model} is employed. For each ($m_1, m_2$) pair, an analysis is initially performed over the user-supplied calibration set, yielding the achieved PSNR and latency for different values of $TV^{\text{thr}}$. As a final step, MobiSR selects the fastest design that lies within the tolerated error of the target application.

\begin{table*}[t]
\centering
\begin{tabular}{l l c || r r r | r r r r }
    \toprule
    \textbf{Model} & & \textbf{Params (K)} & \multicolumn{3}{c|}{\textbf{Latency (ms)}} & \multicolumn{4}{c}{\textbf{Average PSNR/SSIM}$^\dagger$} \\
    & & & \multicolumn{1}{c}{CPU} & \multicolumn{1}{c}{GPU} & DSP & \multicolumn{1}{c}{Set5} & \multicolumn{1}{c}{Set14} & \multicolumn{1}{c}{B100} & Urban100 \\
    \midrule
    SRCNN & \cite{SRCNN} & \phantom{0}\textbf{57} & 9742.97 & \textbf{584.83} & \textbf{656.44} & 30.47/0.8610 & 27.57/0.7528 & 26.89/0.7108 & 24.51/0.7232 \\
    VDSR & \cite{VDSR} & 665 & 198027.52 & 7164.60 & 2623.61 & 31.53/0.8840 & \textbf{28.42/0.7830} & 27.29/0.7262 & 25.18/0.7534 \\
    FEQE-P & \cite{Vu2018FastAE} & \phantom{0}96 & \textbf{2996.92} & 911.61 & 1475.45 & 31.53/0.8824 & 28.21/0.7714 & 27.32/0.7273 & 25.32/0.7583 \\
    $m_{\text{ref}}$ & & 152 & 4570.08 &  2792.43 & 1220.00 & \textbf{31.73/0.8873} & 28.24/0.7729 & \textbf{27.33/0.7283} & \textbf{25.34/0.761} \\

    \bottomrule
    \multicolumn{7}{l}{\small{$\dagger$Calculated using full 32-bit floating-point precision (FP32).}}
\end{tabular}
\caption{Comparison of reference model with state-of-the-art efficient SR models ($\times$4 upscaling).}
\label{tab:compare_soa}
\end{table*}

\section{Evaluation}
\label{sec:eval}

This section presents the effectiveness of MobiSR in significantly improving the performance of on-device super-resolution by examining its core components and comparing with the currently standard implementations and highly optimized difficulty-unaware designs.

\subsection{Experimental Setup}
\label{subsec:exp_setup}

In our experiments, we target Intrinsyc's Open-Q 845 board mounting the Qualcomm Snapdragon 845 SoC (SDM845). SDM845 integrates an octa-core Kryo 385 CPU alongside an Adreno 630 mobile GPU and a Hexagon 685 DSP on the same chip.
\footnote{MobiSR can also target modern mobile chipsets equipped with CPU, GPU and NPU/DSP engines such as Samsung Exynos 9820, Qualcomm Snapdragon 855 and Huawei Kirin 810 SoCs.} All SR models were developed and trained using PyTorch (v1.0) and run on the Open-Q 845 board using the Snapdragon Neural Processing Engine (SNPE)\footnote{https://developer.qualcomm.com/software/qualcomm-neural-processing-sdk} SDK (v1.21). SNPE allows targeting all three CPU, GPU and DSP engines of the SDM845 platform with highly optimized execution of CNN layers. The three compute engines employ different precision for data representation; namely the CPU, GPU and DSP use single-precision floating-point (FP32), half-precision floating-point (FP16) and 8-bit fixed-point (INT8) respectively for both storage and computation. All models that were run on the Hexagon DSP were first quantized offline to INT8 using linear quantization, with the per-layer scaling factors tuned based on the dynamic range of weights and activations on the DIV2K validation set. 

\textbf{Datasets and Training Scheme.}
Following the common practice of the super-resolution community \cite{EDSR, SRMDNF, RDN}, all SR models were trained on the training set of the DIV2K dataset \cite{DIV2K} and validated on its validation set, comprising 800 and 100 images of 2K resolution with diverse contents respectively. For the evaluation, four benchmark datasets were used which constitute the standard for assessing SR models in the super-resolution literature \cite{DIV2K}: Set5 \cite{Set5} and \mbox{Set14 \cite{Set14}} comprising five and fourteen images respectively that are commonly used across the image processing community, \mbox{B100 \cite{B100}} with 100 images of real-life scenes and Urban100 \mbox{\cite{Urban100}} consisting of 100 images depicting urban environments.

For the training of the SR models, we employ a similar scheme to the one used by \cite{RCAN} and \cite{EDSR}. First, data augmentation was applied on the DIV2K training set by randomly flipping horizontally and rotating by $90\degree$, and all images were normalized by subtracting the training set's mean. Next, training was performed in 300 epochs using an Adam optimizer \cite{Kingma_2014} with $\beta_1 = 0.9$, $\beta_2 = 0.999$, $\epsilon = 10^{-8}$ and L1 loss. Each mini-batch consists of 16 RGB patches with input size of 96$\times$96 for both $\times$2 and $\times$4 upscaling. The starting learning rate was set to $10^{-4}$ and was halved after 200 epochs. Lastly, we train $\times$2 models from scratch and use them as pre-trained models to train $\times$4 models, confirming the findings of \cite{EDSR} that using the weights of the $\times$2 models as initial weight values for $\times$4 models leads to faster training convergence.

\textbf{Implementation Details.}
All SR designs presented in this section were run on SDM845 using the high-performance profile of the SNPE SDK that configures the hardware for maximum processing speed. All inputs to the SR models during on-device inference are partitioned into overlapping patches of size $90 \times 160$, with partial results stitched together at the end. The reported latency in all experiments is based on the average across 100 runs, with the latency measurements conducted using the SNPE's timing utilities. The images in the aforementioned SR datasets have different sizes and therefore we report the average latency taken to upscale an image in the given dataset. For all other experiments, we assume a target high-resolution image with 720p resolution ($1280 \times 720$).

\begin{figure}[t]
\centering
\captionsetup[subfigure]{justification=centering}
\begin{subfigure}[b]{0.25\textwidth}
\centering 
\includegraphics[trim={7cm 6cm 9.5cm 5cm},clip,width=1\linewidth]{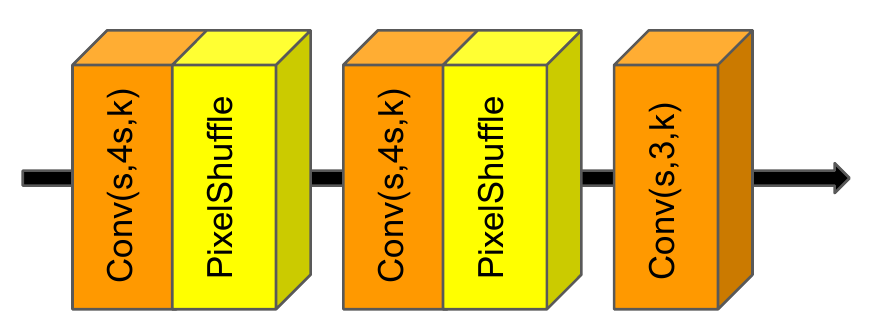}
\caption{RCAN $\times$4 Upsampling Module}
\end{subfigure}
~
\begin{subfigure}[b]{0.25\textwidth}
\centering
\includegraphics[trim={7cm 6cm 9.5cm 5cm},clip,width=1\linewidth]{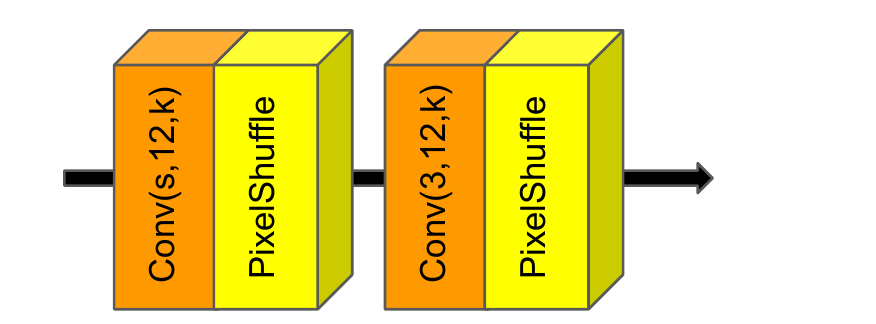}
\caption{$m_{\text{ref}}$ $\times$4 Upsampling \\ Module}
\end{subfigure}
    \caption{We reduce the number of feature maps in RCAN to improve processing speed. $Conv(S,D,K)$ is a $K$$\times$$K$ convolution with $S$ input and $D$ output feature maps.}
\label{fig:upsample_module}
\vspace{0.25cm}
\end{figure}

\begin{table*}[t]
\centering
\begin{tabular}{l l || r r r | r r r | r r }
    \toprule
\textbf{Model} & \textbf{Params (K)} & \multicolumn{3}{c|}{\textbf{Latency (ms)}} & \multicolumn{3}{c|}{\textbf{Speedup}} & \multicolumn{2}{c}{\textbf{Error}} \\
    & & \multicolumn{1}{c}{CPU} & \multicolumn{1}{c}{GPU} & DSP & \multicolumn{1}{c}{CPU} & \multicolumn{1}{c}{GPU} & DSP & \multicolumn{1}{c}{CPU/GPU$^\dagger$} & DSP \\
    \midrule
    $m_{\text{ref}}$ & 152 & 4570.08 &  2792.43 & 1220.0 & 1.00 & 1.00 & 1.00 & 0.00\% & 0.00\% \\
    $m_{\text{rn(r=2)}}$ & 58  
    & 2694.94 & 2626.78 & 1508.56 & 1.69 & 1.06 & 0.80 & 2.84\% & 14.27\% \\
    $m_{\text{rn(r=4)}}$ & 22  
    & 1434.42 & 2657.49 & 1561.97 & 3.18 & 1.05 & 0.78 & 3.94\% & 22.39\% \\
    $m_{\text{rxn}}$& 30 
    & 1850.82 & 10692.13 & 2508.63 & 2.46 & 0.26 & 0.48 & 3.70\% & 19.27\% \\
    $m_{\text{m1}}$& 30 
    & 1969.74 & 2723.66 & 1080.59 & 2.32 & 1.02 & 1.12 & 2.48\% & 4.15\% \\
    $m_{\text{eff}}$& 24
    & 1284.21 & 2700.24 & 1398.407 & 3.55 & 1.03 & 0.87 & 3.19\% & 5.71\% \\
    $m_{\text{m2}}$& 88
    & 4045.80 & 2846.70 & 1327.65 & 1.12 & 0.98 & 0.91 & 2.32\% & 2.11\% \\
    $m_{\text{clc}}$& 30
    & 1910.74 & 2722.86 & 1061.38 & 2.39 & 1.02 & 1.14 & \textbf{1.93\%} & \textbf{0.87\%} \\
    $m_{\text{s1}}$& 13 
    & 1263.90 & 12367.24 & 3060.82 & 3.61 & 0.22 & 0.39 & 4.69\% & 26.1\% \\
    $m_{\text{s2}}$& 17
    & 1023.26 & 2595.59 & 973.07 & \textbf{4.46} & \textbf{1.07} & \textbf{1.25} & 3.03\% & 3.07\% \\
    \midrule
    
    \multicolumn{7}{l}{\small{$\dagger$ We obtained similar results on CPU with FP32 and GPU with FP16.}}
\end{tabular}
\caption{Performance of our explored model space for $\times$4 upscaling. Error drop is based on PSNR on Urban100.}
\label{tab:compression_techniques}
\vspace{-0.2cm}
\end{table*}

\subsection{Evaluation of Model Transformations}
\label{sec:eval_trans}

In MobiSR, the user supplies a reference model and the framework applies a series of model transformations to generate a set of compressed models. This set is then automatically pruned to remove suboptimal models, resulting in a list of Pareto-optimal candidate models to select from in order to produce the resulting two-model SR system. In our experiments, we exemplify this process by selecting a reference model, $m_{\text{ref}}$, that is comparable to the state-of-the-art models in the existing literature for mobile SR and then pass it through MobiSR. 

\textbf{Reference Model Selection.}
We adopted the residual channel attention network (RCAN) \cite{RCAN} as our reference model, $m_{\text{ref}}$, as RCAN yields the state-of-the-art performance based on PSNR/SSIM among large-scale SR models. In order for RCAN to be comparable to existing state-of-the-art mobile SR models, its architecture was modified by reducing the number of residual groups to $3$, the number of residual channel attention blocks to $10$, and the number of feature maps to $16$. Additionally, to further reduce the computational cost of the reference model, the number of feature maps in the upscaling module was reduced by a factor of $5$ and the last convolutional layer was removed. Fig. \ref{fig:upsample_module} shows the slight change in the upsampling module between RCAN and $m_{\text{ref}}$.

\begin{figure}[t] 
\centering
\includegraphics[trim={3cm 0cm 3cm 2cm},clip,width=0.5\textwidth]{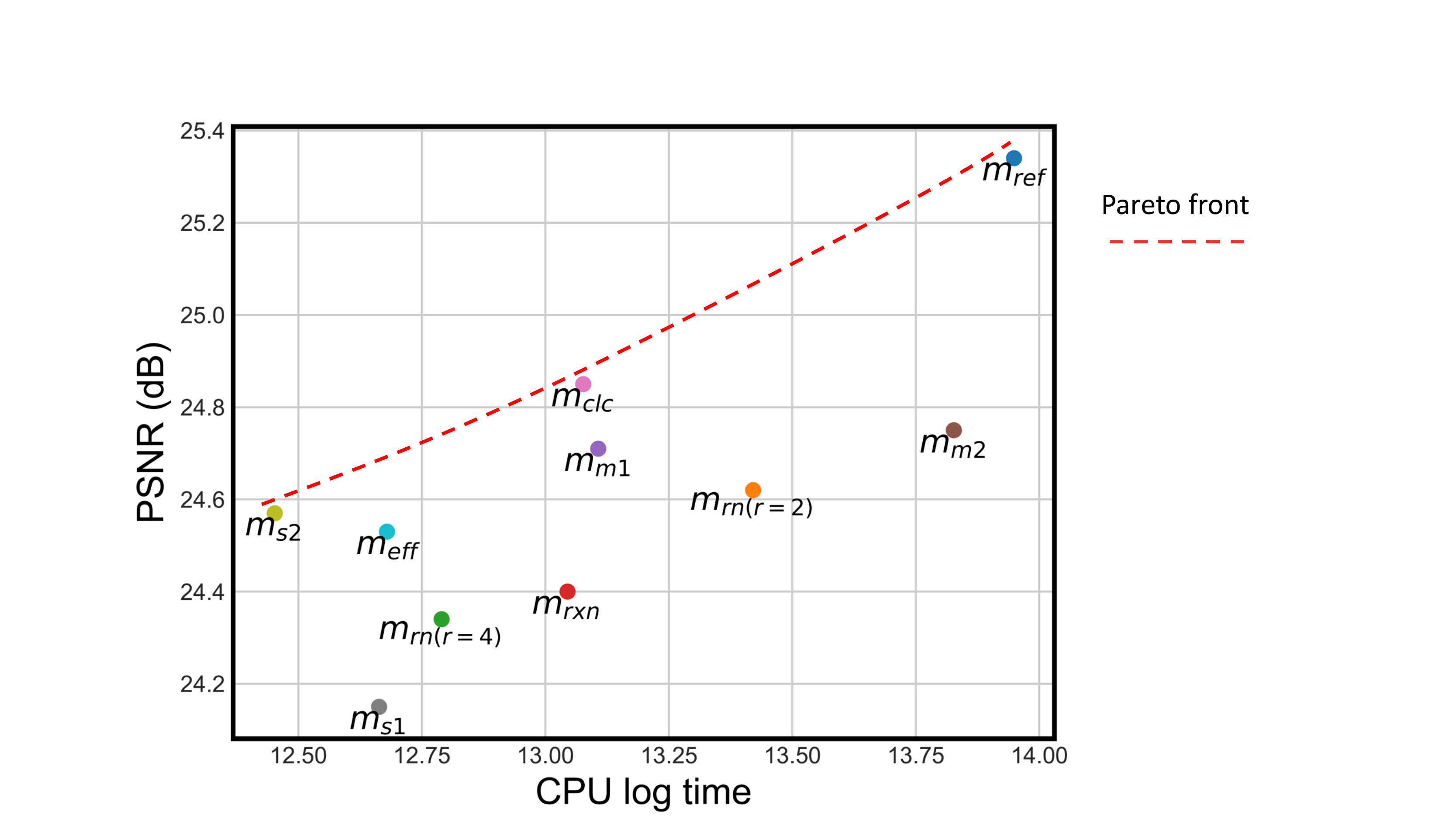}
\caption{PSNR vs CPU latency of MobiSR-generated models on SDM845 ($\times$4 upscaling on Urban100).}
\label{fig:psnr_cpu}
\vspace{0.2cm}
\end{figure}

As shown on Table \ref{tab:compare_soa}, by constructing a shallower variant of RCAN, we are able to achieve comparable results with state-of-the-art SR models that are hand-optimized for increased efficiency. Notably, our reference model manages to outperform the winning model of the 2018 PIRM Challenge \cite{Ignatov_2018} on perceptual SR on mobile, FEQE, by $20\%$ when run on the Hexagon DSP and achieves higher PSNR across all four SR datasets. Furthermore, $m_{\text{ref}}$ yields an average speedup of 16.01$\times$ (6.2$\times$ geo. mean) over VDSR with 4.3$\times$ fewer parameters and achieves an average PSNR improvement of 0.8 dB over the lightweight SRCNN. Regardless, MobiSR accepts any starting reference model and searches for a set of model transformations that will work best for that reference model on the given compute engines.

\begin{figure}[t]
\centering
\includegraphics[trim={3cm 0cm 3cm 2cm},clip,width=0.5\textwidth]{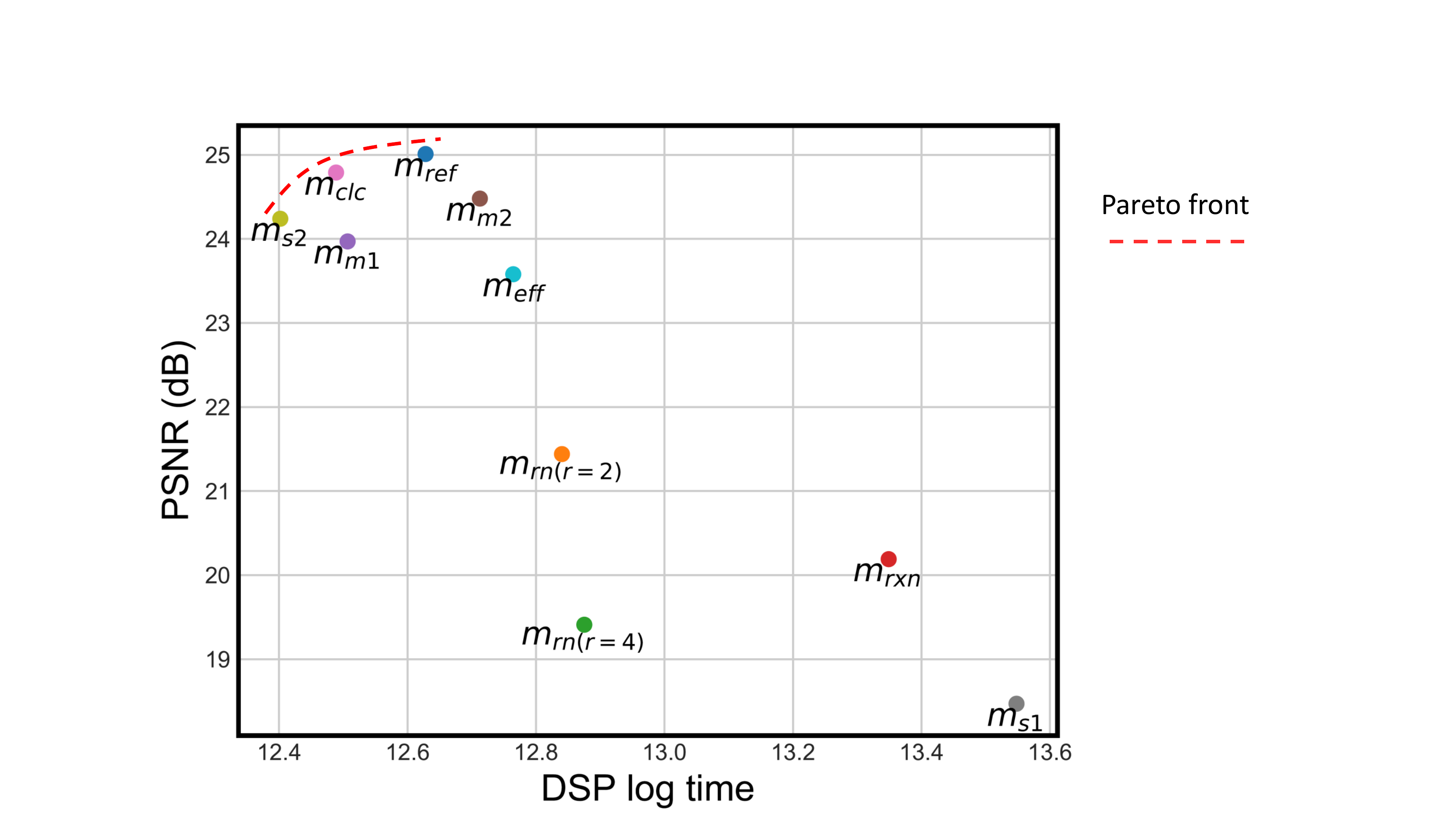}
\caption{PSNR vs DSP latency of MobiSR-generated models on SDM845 ($\times$4 upscaling on Urban100).}
\label{fig:psnr_dsp}
\vspace{0.2cm}
\end{figure}

\begin{figure*}[t]
\centering
\includegraphics[trim={0cm 0cm 0cm 0cm},clip,width=1\textwidth]{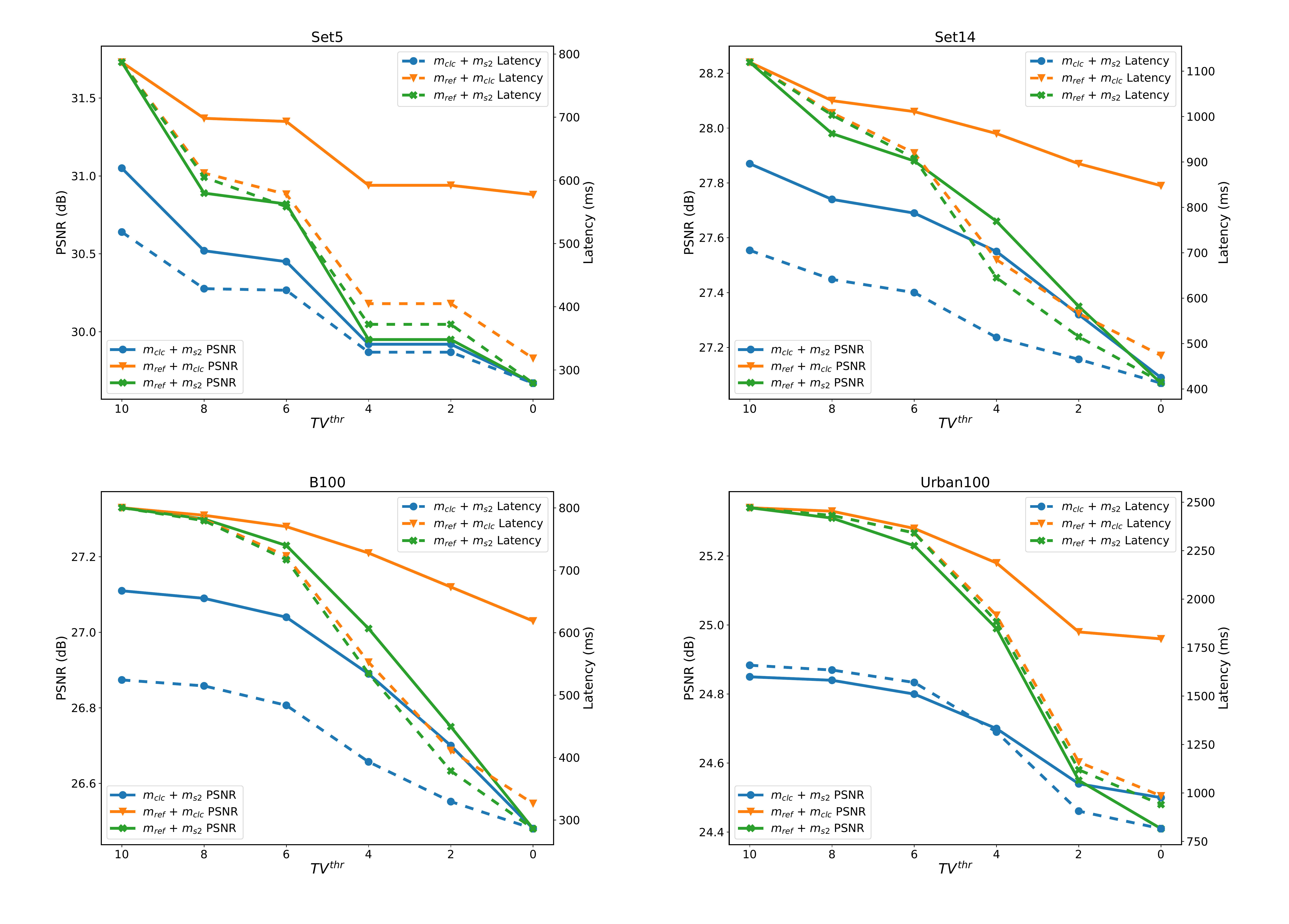}
\vspace{-1.0cm}
\caption{Achieved PSNR and measured performance as a function of TV on SDM854.}
\label{fig:psnr_latency_tv}
\end{figure*}

\textbf{Explored Model Space.}
Based on the findings of recently proposed high- and low-level vision models, we examined specific transformation combinations from the transformation set $T$ (detailed in Section \ref{subsec:model_space}) on our reference model by focusing on the ones that have demonstrated the highest effectiveness in the deep learning literature. Table \ref{tab:compress_tech} details the topologies of the MobiSR-generated compressed models together with the associated transformations that were applied over our reference model. Specifically, given the reference model, MobiSR replaces all 3$\times$3 convolutional layers that lie in the core of the network, excluding the first layer and those within the upsampling block, using a subset of transformations. After this step, the obtained compressed models are retrained from scratch following the training scheme of Section \ref{subsec:exp_setup}.

Table \ref{tab:compression_techniques} lists the attainable latency of each model, the speedup over the reference model and the error with respect to PSNR, as obtained by MobiSR's \textit{On-device} and \textit{Image Quality Evaluator} modules. Apart from the latency-PSNR trade-off of the generated compressed models, Table \ref{tab:compression_techniques} also highlights the compatibility of each compute engine with the various transformations that have been applied. Different compute engines are more highly optimized for a different subset of transformations. For instance, the sole use of bottleneck residual blocks, $m_\text{rn(r=4)}$, obtains a greater speedup on the CPU, but lower gains on the DSP as compared to the sole use of depthwise separable convolutions, $m_\text{m1}$. Furthermore, models that employ group convolutions, such as $m_\text{s1}$ and $m_\text{rxn}$, yield worse latency when executed on the GPU due to suboptimal mapping. Additionally, the 8-bit quantization of the DSP had a severe impact on the representational capacity of compressed models that utilized the residual bottleneck blocks, such as $m_\text{rn(r=2)}$, $m_\text{rn(r=4)}$ and $m_\text{rxn}$. As a result, selecting the highest performing set of compressed models is dependent on both the provided reference model and the available compute engines.

Overall, Fig. \ref{fig:psnr_cpu} and \ref{fig:psnr_dsp} depict the PSNR-latency space of the generated models on the CPU and DSP of the Qualcomm SDM845 respectively. In this case, the framework picked the same three models, namely $m_{\text{ref}}$, $m_{\text{clc}}$ and $m_{\text{s2}}$, that lie on the Pareto fronts of all three compute engines.

\setlength{\tabcolsep}{2pt}
\begin{table*}[t]
\centering
\begin{tabular}{l || l l | l  l | l l | l l }
    \toprule
    \textbf{Model} & \multicolumn{2}{c|}{\textbf{Set5}} & \multicolumn{2}{c|}{\textbf{Set14}} & \multicolumn{2}{c|}{\textbf{B100}} & \multicolumn{2}{c}{\textbf{Urban100}}  \\
    \textbf{Pair} & \multicolumn{1}{c}{Speedup} & \multicolumn{1}{l|}{Avg/G. Mean} & \multicolumn{1}{c}{Speedup} & \multicolumn{1}{l|}{Avg/G. Mean} & \multicolumn{1}{c}{Speedup} & \multicolumn{1}{l|}{Avg/G. Mean} & \multicolumn{1}{c}{Speedup} & \multicolumn{1}{l}{Avg/G. Mean} \\
    
    \midrule

    \multicolumn{9}{c}{\textbf{Running $m_{\text{ref}}$ on the CPU}} \\
    \hline
    ($m_{\text{ref}}$, $m_{\text{clc}}$) & 1.74$\times$-4.31$\times$ & 2.90$\times$/2.78$\times$ & 1.96$\times$-4.65$\times$ & 3.05$\times$/2.90$\times$ &  1.76$\times$-4.31$\times$ & 2.63$\times$/2.47$\times$ &  2.35$\times$-5.91$\times$ & 3.53$\times$/3.28$\times$  \\
    
    ($m_{\text{ref}}$, $m_{\text{s2}}$) & 1.74$\times$-4.91$\times$ & 3.12$\times$/2.94$\times$ &  1.97$\times$-5.30$\times$ & 3.26$\times$/3.05$\times$ &  1.76$\times$-4.91$\times$ & 2.79$\times$/2.58$\times$ &  2.35$\times$-6.18$\times$ & 3.62$\times$/3.34$\times$ \\
    
    ($m_{\text{clc}}$, $m_{\text{s2}}$) & 2.64$\times$-4.91$\times$ & 3.72$\times$/3.64$\times$ & 3.12$\times$-5.34$\times$ & 4.09$\times$/4.01$\times$ & 2.68$\times$-4.91$\times$ & 3.51$\times$/3.42$\times$ & 3.51$\times$-7.13$\times$ & 4.79$\times$/4.59$\times$ \\
    \hline 
    \multicolumn{9}{c}{\textbf{Running $m_{\text{ref}}$ on the GPU}} \\
    \hline
    ($m_{\text{ref}}$, $m_{\text{clc}}$) & 1.06$\times$-2.63$\times$ & 1.78$\times$/1.70$\times$ & 1.20$\times$-2.84$\times$ & 1.86$\times$/1.77$\times$ &  1.07$\times$-2.63$\times$ & 1.61$\times$/1.51$\times$ &  1.44$\times$-3.61$\times$ & 2.16$\times$/2.01$\times$  \\
    
    ($m_{\text{ref}}$, $m_{\text{s2}}$) & 1.06$\times$-3.00$\times$ & 1.91$\times$/1.80$\times$ &  1.20$\times$-3.24$\times$ & 1.99$\times$/1.87$\times$ &  1.07$\times$-3.00$\times$ & 1.71$\times$/1.59$\times$ &  1.44$\times$-3.78$\times$ & 2.21$\times$/2.04$\times$ \\
    
    ($m_{\text{clc}}$, $m_{\text{s2}}$) & 1.61$\times$-3.00$\times$ & 2.27$\times$/2.23$\times$ & 1.91$\times$-3.26$\times$ & 2.50$\times$/2.45$\times$ & 1.64$\times$-3.00$\times$ & 2.15$\times$/2.09$\times$ & 2.14$\times$-4.36$\times$ & 2.93$\times$/2.80$\times$ \\
    \hline
    \multicolumn{9}{c}{\textbf{Running $m_{\text{ref}}$ on the CPU \& GPU}} \\
    \hline
    ($m_{\text{ref}}$, $m_{\text{clc}}$) & 1.28$\times$-2.47$\times$ & 1.80$\times$/1.75$\times$ & 1.11$\times$-2.36$\times$ & 1.66$\times$/1.59$\times$ &  1.02$\times$-2.45$\times$ & 1.59$\times$/1.51$\times$ &  1.01$\times$-2.51$\times$ & 1.60$\times$/1.49$\times$  \\
    
    ($m_{\text{ref}}$, $m_{\text{s2}}$) & 1.30$\times$-2.82$\times$ & 1.95$\times$/1.87$\times$ &  1.11$\times$-2.69$\times$ & 1.79$\times$/1.69$\times$ &  1.02$\times$-2.79$\times$ & 1.71$\times$/1.59$\times$ &  1.01$\times$-2.62$\times$ & 1.64$\times$/1.52$\times$ \\
    
    ($m_{\text{clc}}$, $m_{\text{s2}}$) & 1.52$\times$-2.82$\times$ & 2.13$\times$/2.09$\times$ & 1.58$\times$-2.71$\times$ & 2.08$\times$/2.04$\times$ & 1.52$\times$-2.79$\times$ & 2.09$\times$/2.04$\times$ & 1.49$\times$-3.03$\times$ & 2.04$\times$/1.95$\times$ \\

    \bottomrule
\end{tabular}
\vspace{0.2cm}
\caption{Performance comparison of Pareto-optimal model pairs with faithful reference model $m_{\text{ref}}$.}
\label{tab:single_spd}
\vspace{-0.5cm}
\end{table*}

\subsection{MobiSR PSNR and Performance vs TV}

In this section, the PSNR and performance of the MobiSR-generated designs are evaluated as a function of total-variation threshold. Given the three Pareto-optimal models from Section \ref{sec:eval_trans} and the pruning strategy that dictates that $m_2$ is more compact than $m_1$, three model pairs were selected in the valid design space; namely ($m_{\text{ref}}$, $m_{\text{clc}}$), ($m_{\text{ref}}$, $m_{\text{s2}}$) and ($m_{\text{clc}}$, $m_{\text{s2}}$). 

Fig. \ref{fig:psnr_latency_tv} shows the measured latency on SDM854 and the achieved PSNR across different TV thresholds for the four SR datasets. When $TV^{\text{thr}}$ has substantially high values (towards the left hand side of the plots), the majority of incoming samples is processed by $m_1$ on the CPU and GPU of SDM845. In this manner, PSNR remains high, but at the cost of increased latency due to the underutilization of the DSP. As $TV^{\text{thr}}$ decreases from left to right, the three model pairs trade off a decreased PSNR for substantially reduced processing latency. Eventually, as $TV^{\text{thr}}$ reaches very low values, the DEU relaxes the constraints and its scheduling policy reduces to a load balancing of the incoming samples across the three compute engines, without the need to exploit the upscaling difficulty of each sample. In this manner, the highest speed up is achieved for low $TV^{\text{thr}}$ at the expense of a significant drop in the achieved PSNR. 

\begin{figure*}[t]
\centering
\includegraphics[trim={0cm 0cm 0cm 0cm},clip,width=1\textwidth]{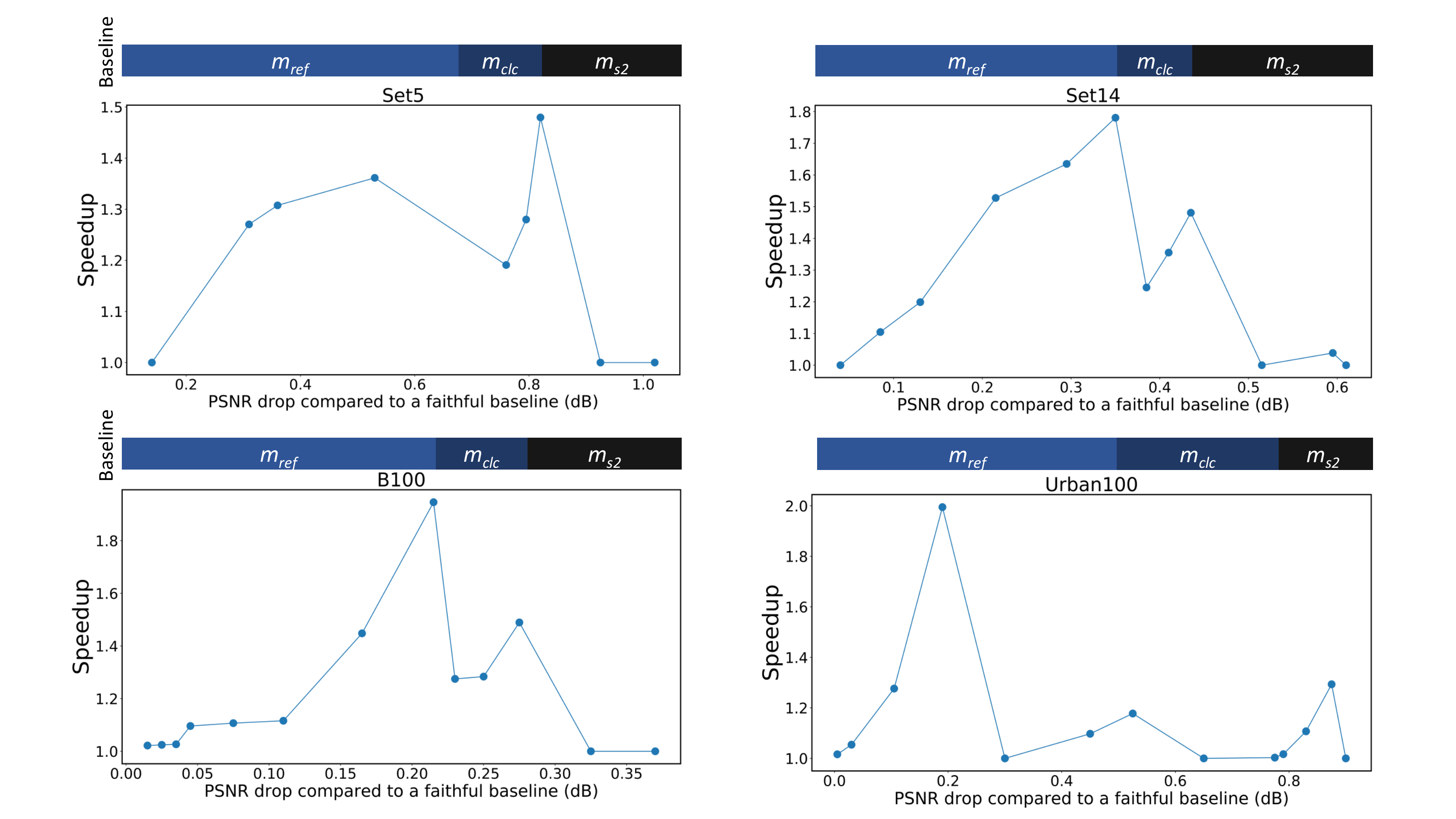}
\caption{MobiSR's speedup as a function of error degradation.}
\label{fig:speedup}
\end{figure*}

Table \ref{tab:single_spd} lists the speedup gains of the three model pairs over the execution of $m_{\text{ref}}$ on the CPU, GPU, and both CPU and GPU. Since $m_{\text{ref}}$ is mapped only on the CPU and/or GPU, no degradation of PSNR is induced due to the 8-bit operations of the DSP. Overall, the parametrization of the DEU based on $TV^{\text{thr}}$ allows the tuning of the system at a fine granularity so that even a small increase in the application-level error tolerance can be capitalized as reduced processing latency.

\begin{table*}[t!]
\vspace{-1cm}
\centering
\begin{tabular}{l || r | r | l}
    \toprule
    \textbf{Model} & \textbf{$T^*$} & \textbf{Inspired By} & \textbf{Building Block} \\
\midrule
    $m_{\text{rn}}$ & $\{rb(2)\}$ \& $\{rb(4)\}$ & ResNet \cite{He_2016} & 
    \pbox{20cm}{\vspace{0.1cm}
    $y\leftarrow Conv(x,s,\frac{S}{r},(1,1),1)$ \\ 
 $y\leftarrow ReLU(y)$ \\
     $y\leftarrow Conv(y,\frac{S}{r},\frac{S}{r},(K_h,K_w),1)$ \\
     $y\leftarrow ReLU(y)$ \\
     $y\leftarrow Conv(y,\frac{S}{r},s,(1,1),1)$ \\
     $y\leftarrow y + x$
     \vspace{0.1cm}}  \\\hline

    $m_{\text{rxn}}$ & $\{rb(2),grp(4)\}$ & ResNeXt \cite{Xie_2017} & 
    \pbox{20cm}{\vspace{0.1cm}
    $y\leftarrow Conv(x,S,\frac{S}{r},(1,1),1)$ \\ 
     $y\leftarrow ReLU(y)$ \\
     $y\leftarrow Conv(y,\frac{S}{r},\frac{S}{r},(K_h,K_w),g)$ \\
     $y\leftarrow ReLU(y)$ \\
     $y\leftarrow Conv(y,\frac{S}{r},S,(1,1),1)$ \\
     $y\leftarrow y + x$
     \vspace{0.1cm}}  
     \\\hline

    $m_{\text{m1}}$ & $\{dpth\}$ & MobileNet \cite{mobilenetv1} &
    \pbox{20cm}{\vspace{0.1cm}
    $y\leftarrow Conv(x,S,S,(K_h,K_w),s)$ \\ 
     $y\leftarrow ReLU(y)$ \\
     $y\leftarrow Conv(y,S,D,(1,1),1)$
     \vspace{0.1cm}} \\\hline

    $m_{\text{eff}}$ & $\{rb(2),dpth,sep\}$ & EffNet \cite{effnet} &
    \pbox{20cm}{\vspace{0.1cm}
    $y\leftarrow Conv(x,S,\frac{S}{r},(1,1),1)$ \\ 
     $y\leftarrow ReLU(y)$ \\
     $y\leftarrow Conv(y,\frac{S}{r},\frac{S}{r},(1,K_w),\frac{S}{r})$ \\
     $y\leftarrow Conv(y,\frac{S}{r},\frac{S}{r},(K_h,1),\frac{S}{r})$ \\
     $y\leftarrow ReLU(y)$ \\
     $y\leftarrow Conv(y,\frac{S}{r},S,(1,1),1)$ \\
     $y\leftarrow y + x$
     \vspace{0.1cm}}  
     \\\hline
     
    $m_{\text{m2}}$ & $\{dpth,invr(2)\}$ & MobileNetV2 \cite{mobilenetv2} & 
    \pbox{20cm}{\vspace{0.1cm}
    $y\leftarrow Conv(x,S,S \times e,(1,1),1)$ \\ 
     $y\leftarrow ReLU(y)$ \\
     $y\leftarrow Conv(y,S \times e,S \times e,(K_h,K_w),S \times e)$ \\
     $y\leftarrow ReLU(y)$ \\
     $y\leftarrow Conv(y,s \times e,S,(1,1),1)$ \\
     $y\leftarrow y + x$
     \vspace{0.1cm}}
    
    \\\hline
    $m_{\text{clc}}$ & $\{grp(16),chlshf\}$ & ClcNet \cite{clcnet} & 
    \pbox{20cm}{\vspace{0.1cm}
    $y\leftarrow Conv(x,S,S,(K_h,K_w),g)$ \\ 
     $y\leftarrow chlshf(y)$ \\
     $y\leftarrow Conv(y,S,D,(1,1),1)$
     \vspace{0.1cm}}
     
     \\\hline
    $m_{\text{s1}}$ & $\{rb(2),grp(4),dpth,chlshf\}$ & ShuffleNet \cite{shufflenetv1} & 
    \pbox{20cm}{\vspace{0.1cm}
    $y\leftarrow Conv(x,S,\frac{S}{r},(1,1),g)$ \\ 
     $y\leftarrow ReLU(y)$ \\
     $y\leftarrow chlshf(y)$ \\
     $y\leftarrow Conv(y,\frac{S}{r},\frac{S}{r},(K_h,K_w),\frac{S}{r})$ \\
     $y\leftarrow Conv(y,\frac{S}{r},S,(1,1),g)$ \\
     $y\leftarrow y + x$
     \vspace{0.1cm}}  
     
     \\\hline
    $m_{\text{s2}}$ & $\{chlsplt,dpth,chlshf\}$ & ShuffleNet V2 \cite{shufflenetv2} & 
    \pbox{20cm}{\vspace{0.1cm}
    $y_1, y_2\leftarrow chlsplit(x)$ \\
    $y_1\leftarrow Conv(y_1,\frac{S}{2},\frac{S}{2},(1,1),1)$ \\ 
     $y_1\leftarrow ReLU(y_1)$ \\
     $y_1\leftarrow Conv(y_1,\frac{S}{2},\frac{S}{2},(K_h,K_w),\frac{S}{2})$ \\
     $y_1\leftarrow Conv(y_1,\frac{S}{2},\frac{S}{2},(1,1),1)$ \\
     $y\leftarrow [y_1, y_2]$ \\
     $y\leftarrow chlshf(y)$
     \vspace{0.1cm}}  
    
    \\\hline
    \midrule

    \bottomrule
\end{tabular}
\caption{\small All 3$\times$3 convolutional layers in $m_{\text{ref}}$ are replaced with the corresponding building blocks where $ReLU$ represents the rectifier activation function, and $Conv(x, S,D,(K_h,K_w),g)$ represents a $K_h$$\times$$K_w$convolutional layer with input $x$, $S$ input channels, $D$ output channels and $g$ groups.}
\label{tab:compress_tech}
\end{table*}

\subsection{Evaluation of MobiSR Performance}
\label{subsec:mobisr_speedup}
This section presents the performance gains of MobiSR with respect to processing speed. 
This is investigated by comparing the generated two-model design for different PSNR drop values with a baseline single-model network. 
For each interval of PSNR drop, each MobiSR instance is compared with the fastest baseline single-model architecture that achieves the same or higher PSNR as the MobiSR system (shown on top of each plot in Fig. \ref{fig:speedup}). 
The single-model baselines do not employ MobiSR's TV-based scheduling; instead each model is allowed to run in one of two modes: \textit{i)} either with load balancing across the CPU and GPU and no PSNR degradation or \textit{ii)} with load balancing across CPU, GPU, and DSP with PSNR drop due to the DSP's reduced precision. 
In this respect, the fastest single model that satisfies the PSNR drop constraint is selected at each PSNR drop interval. The overall measured runtime includes the DEU, processing all patches and the overhead of combining the partial results to construct the final high-resolution image.

Fig. \ref{fig:speedup} presents the achieved speedup across a wide range of PSNR tolerance values on the SDM845 platform when targeting the four benchmark datasets. When minimal to no PSNR drop is allowed (towards the left of Fig. \ref{fig:speedup}), MobiSR selects a strict scheduling policy for the DEU with high TV values. In this manner, the large majority of patches are processed by $m_1$ on the CPU and GPU and the DSP remains underutilized, leading to minimal speedup. As more error is allowed, the proposed system outperforms the baseline by up to 47\%, 78\%, 94\% and 29\% for the same PSNR drop budget in Set5, Set14, B100 and Urban100 respectively. 

Finally, in the case of high error tolerance, the speedup becomes less significant as uninformed load balancing using the fastest compressed model across the CPU, GPU, and DSP becomes the fastest design.

\section{Related Work}
\label{sec:related work}

The emergence of mobile image-centric applications has attracted the attention of the computer vision community, with efforts for alleviating the large compute demands of large-scale SR models. Addressing on-device SR from a model perspective, SRCNN \cite{SRCNN} and the second-generation \mbox{FSRCNN \cite{FSRCNN}} were first proposed as efficient neural architectures for SR, consisting of only three convolutional layers. By aiming to improve the image quality, the VDSR \cite{VDSR} network employed a deeper design consisting of twenty layers. With a direction towards mobile settings, CARN-M \cite{CARN} was proposed as a lightweight variant of the CARN architecture. Inspired by MobileNet \cite{mobilenetv1}, CARN-M employs recursive blocks and group convolutions to reduce the storage and compute requirements at inference time. In 2018, FEQE \cite{Vu2018FastAE} introduced the desubpixel block, enabling a lossless downsampling at the start of the network, while reducing the computation cost throughout the rest of the network. The aforementioned works are primarily hand-tuned architectures with manually selected architectural choices aiming to reach a balance between image quality and computation cost. In this paper, we focus on the largely unexplored space of applying model transformations in a hardware-aware manner with the goal to tailor the generated system to both the application-level image quality requirements and the target mobile platform characteristics.

From a systems perspective, apart from task-agnostic frameworks for executing deep neural networks on mobile platforms \cite{Huynh_2016,Oskouei_2016,deepx_2016,caffepresso_2017,Song_2018}, research efforts have mainly focused in the direction of 1) cascade systems \cite{mcdnn_2016,videostorm_2017,noscope_2017,focus_2018,shen_2017,cascadecnn_2018}, 2) early-exit classifiers \cite{branchynet_2016,msdnet_2018,overthink_2019} and 3) specialized accelerators for CNNs \cite{Qiu_2016,Venieris_2018} and SR models \cite{He_2018,fpga_sr_2018}.

\textbf{Cascade systems.}
Cascade systems base their operation on conditionally passing input samples through a pipeline of classifiers based on information obtained at each classification stage. \texttt{VideoStorm} \cite{videostorm_2017}, \texttt{NoScope} \cite{noscope_2017}, \texttt{Focus} \cite{focus_2018} and Shen \textit{et al.} \cite{shen_2017} focus on the task of issuing queries on video databases. A common element between these systems and MobiSR is the use of multiple networks. However, a key differentiating factor in the model generation approach is that, by exploiting video-specific optimization opportunities, these systems train class-specialized models based on the object classes that appear more often in a given video stream. In contrast, the generative nature of the super-resolution task is not amenable to such an approach. Furthermore, at the run-time model selection stage, each stage of the cascade determines whether a particular object class is present or not, and if not, the input sample is propagated to the next classification stage. Contrary to this approach, MobiSR's DEU determines which model to use based on the input image complexity and the current load of the available compute engines, without requiring information to be passed between models.

In a similar manner to MobiSR, MCDNN \cite{mcdnn_2016} employs a form of model selection. However, while MCDNN focuses on run-time model selection and the partitioning of computation between cloud and device, MobiSR employs a difficulty-aware mechanism to exploit the heterogeneous compute engines that are available on-device and parallelizes both within an image (\textit{i.e.} parallel processing of patches) and across images (\textit{i.e.} pipelined execution as long as images are available). Moreover, while MCDNN aims to maximize the average accuracy of classification tasks, MobiSR sets a constraint on the PSNR drop and guarantees that the average PSNR will not be compromised below user-specified bounds.

From a target platform perspective, the aforementioned systems are optimized for cloud setups that have substantially different characteristics compared to MobiSR's fully on-device system. In our case, the available compute engines share the same main memory, and in turn the same storage and bandwidth. This poses a significant additional challenge and calls for the mobile-specific methodology of MobiSR to develop and implement high-performing mobile designs. 

Finally, the cascading approach of CascadeCNN \cite{cascadecnn_2018} involves a two-model cascade with each classifier quantized at a different precision level. In this case, input samples are first processed rapidly by an aggressively quantized model. If the prediction confidence of a sample is below a tunable threshold, the input sample is passed to a higher-precision model for recomputation. Despite the fact that variable precision quantization could be integrated in the transformations set of MobiSR, CascadeCNN has so far been evaluated on FPGA-based platforms targeting image recognition tasks.

\textbf{Early-exit classifiers.}
Designs such as BranchyNet \cite{branchynet_2016}, MSDNet \cite{msdnet_2018} and Shallow-Deep Networks \cite{overthink_2019} approach inference acceleration from an architectural aspect. First, they focus on classification rather than generative tasks. Secondly, they explicitly introduce early-exit outputs on a single model in order to reduce the workload-quality characteristics. Nevertheless, by exploiting the fact that different samples require different amount of computation to yield a correct classification, such designs share a similar philosophy to our upscaling-difficulty-aware scheme. However, the criterion to capture an input sample's difficulty is the prediction confidence at each early exit, which is relevant to classification tasks and differs to the upscaling-difficulty metric used by MobiSR for run-time model selection.

\textbf{Hardware acceleration.} Several works have explored the design of custom hardware architectures for the efficient execution of CNN workloads in resource- and power-constrained settings \cite{Qiu_2016, Venieris_2018}. With a focus on SR, \mbox{He \textit{et al.} \cite{He_2018}} proposed a highly optimized FPGA-based hardware accelerator tailored to the FSRCNN \cite{FSRCNN} network. Furthermore, by adopting a hardware-software codesign methodology, Kim \textit{et al.} \cite{fpga_sr_2018} derived a CNN-based SR model and implemented it on an FPGA-based platform. Our work focuses on programmable mobile platforms which are more flexible and enable the efficient execution of SR models in a network-agnostic manner.

\vspace{-0.2cm}
\section{Conclusion}
\label{sec:conclusion}
The MobiSR framework described in this paper uses several techniques to achieve high performance for fully on-device super-resolution. Through the generation of a two-model processing system tailored to the available compute engines of the target mobile platform, the proposed framework demonstrates significant speedup compared to single-model designs without penalizing the achieved image quality. 
By considering the user-specified error tolerance in the design space exploration phase and exploiting the heterogeneous compute engines of commodity mobile platforms, MobiSR is able to deliver high-speed SR on-device while meeting the application-level image quality requirements.

Furthermore, as the proposed methodology is parametrized to target any arbitrary mobile SoC with heterogeneous compute engines, using MobiSR to take advantage of newer emerging platforms that consist of neural accelerators
can be a key enabler for efficient mobile super-resolution with potentially larger room for performance gains.

\balance
\bibliographystyle{ACM-Reference-Format}
\bibliography{sample-base}

\end{document}